\documentclass[10pt,journal,compsoc]{IEEEtran}

\usepackage{xcolor}
\usepackage{tikz}
\usepackage{bm}
\usetikzlibrary{arrows.meta, shapes.misc, shapes}
\usepackage[caption=false]{subfig}

\ifCLASSOPTIONcompsoc
  \usepackage[nocompress]{cite}
\else
  \usepackage{cite}
\fi
\ifCLASSINFOpdf
  \usepackage{graphicx}
\else
\fi
\usepackage{amsmath}
\usepackage{amssymb}
\usepackage{url}
\begin{document}
%
\title{Time Series Forecasting Using LSTM Networks: A~Symbolic Approach}
%
%
%
%

\newcommand{\todo}[1]{{\color{red}#1}}

\author{Steven Elsworth and 
        Stefan G\"{u}ttel
\IEEEcompsocitemizethanks{\IEEEcompsocthanksitem Department of Mathematics, The University of Manchester, Alan Turing Building, Oxford Road, M13\,9PL Manchester, United Kingdom. Email addresses: \texttt{steven.elsworth@manchester.ac.uk}, \texttt{stefan.guettel@manchester.ac.uk}}
\thanks{}}

%
%


\markboth{}%
{}

%



\IEEEtitleabstractindextext{%
\begin{abstract}
Machine learning methods trained on raw numerical time series data exhibit  fundamental limitations such as a high sensitivity to the hyper parameters and even to the  initialization of random weights. A combination of a recurrent neural network with a dimension-reducing symbolic representation is proposed and applied for the purpose of time series forecasting. It is shown that the symbolic representation can help to alleviate some of the aforementioned problems and, in addition, might allow for faster training without sacrificing the forecast performance.
\end{abstract}

}

\maketitle

\IEEEdisplaynontitleabstractindextext

%
\IEEEpeerreviewmaketitle

\ifCLASSOPTIONcompsoc
\IEEEraisesectionheading{\section{Introduction}\label{sec:introduction}}
\else
\section{Introduction}
\label{sec:int}
\fi

%
%
%
%
\IEEEPARstart{T}{ime} series are a common data type occurring in many areas and applications such as finance, supply and demand prediction, and health monitoring. Given a vector of historical time series values $T = [t_1, t_2, \ldots, t_N] \in \mathbb{R}^N$, a prevalent task in time series analysis is to forecast (or ``extrapolate'') future values $\hat{t}_{N+1}, \hat{t}_{N+2}, \ldots$ based on the historical data. 

Time series forecasting methods can be roughly grouped into two main categories: \emph{traditional statistical methods} and methods based on \emph{machine learning models}. While recurrent neural networks (RNNs), which fall into the the latter category, are frequently employed for anomaly detection \cite{MVS15, TLJ16}, classification \cite{CPC18, KMD17, HS03} and forecasting \cite{GLT01, D08, HXX04, BYR17} of time series, a systematic comparison on the M3 Competition\footnote{\url{https://forecasters.org/resources/time-series-data/m3-competition/}} showed that they can be outperformed by traditional statistical methods~\cite{MSA18}. More recently, a hybrid algorithm combining exponential smoothing (a classical statistical method) and recurrent neural networks (machine learning model) called an ES-RNN model \cite{SRP18} won the M4 competition\footnote{\url{https://forecasters.org/resources/time-series-data/m4-competition/}} \cite{SEV20}. It is probably fair to say that, as of now, there is no reliable ``black box'' time series forecasting method available that can achieve human-like performance without some manual pre-processing of the time series data and intensive parameter tuning. It is not even clear what ``reliable'' should mean in this context, given alone the large number of available measures of forecast accuracy \cite{A85,HK06}. 

As we will  demonstrate in this paper, machine learning forecasting methods based on the raw time series values  $t_i$ have some fundamental limitations and drawbacks, such as computationally demanding training phases, a large number of hyper parameters, and even a high sensitivity on  the initialization of random weights. We will show that a dimension-reducing symbolic representation of the time series can significantly speed up the training phase and reduce the model's sensitivity to the hyper parameters and initial weights. The key contributions and outline of this paper are as follows:
\begin{itemize}
\item[--] In Section~\ref{sec:sym} we briefly review the ABBA symbolic representation for time series \cite{EG19b}, and extend it with a new patching procedure that mimics the historical time series data more closely and is visually more appealing.
\item[--] In Section~\ref{sec:lstm} we review the literature standard LSTM and explain how a network is built using LSTM cells. This section serves the purpose of introducing our notation and formalising the LSTM concept. We hope that this section may also serve as a gentle introduction to LSTMs for some readers, similar to the review paper~\cite{HH19} which does not cover recurrent neural networks. 
\item[--] In Section~\ref{sec:tsp} we explain how to build a training set for an RNN model from a single time series. 
Different from other neural network applications, constructing the training data for time series forecasting requires the choice of a lag parameter which can directly affect  the forecasting performance. Furthermore, we explain in detail the differences  between `stateful' and `stateless' training.
\item[--] In Section~\ref{sec:numvssym} we illustrate key differences between LSTM networks  using raw numeric data and the proposed ABBA-LSTM combination. We find that the use of the ABBA representation reduces the network's sensitivity to hyper parameter, reduces the need for linear trend removal, and can lead to forecasts that resemble the behaviour of the historical data more faithfully. 
\item[--] In Section~\ref{sec:rwe} we compare the raw LSTM and ABBA-LSTM approaches on a collection of time series, and find that ABBA-LSTM models are more easily trained while achieving similar forecast performance. We conclude in Section~\ref{sec:con} with a discussion of potential future work.
\end{itemize}

All computational results and figures contained in this paper can be reproduced using the Python codes at
\begin{center}
\url{https://github.com/nla-group/ABBA-LSTM}.
\end{center}
We have used both Keras \cite{KERAS} and Pytorch \cite{PyTorch} for implementing the LSTM networks.

\section{Symbolic representation} \label{sec:sym}
Symbolic representations of time series have become increasingly popular in the data mining community. They have shown to be useful in a variety of applications including classification, clustering, motif discovery and anomaly detection. The key idea is to convert the numerical time series $T = [t_1, t_2, \ldots, t_N]$ into a sequence of symbols $S = [s_1, s_2, \ldots, s_m]$ where each symbol $s_i$ is an element of a finite alphabet $\mathbb{A} = \{a_1, a_2, \ldots, a_k\}$.  

\begin{figure}[h!t]
\includegraphics[width=3.6in]{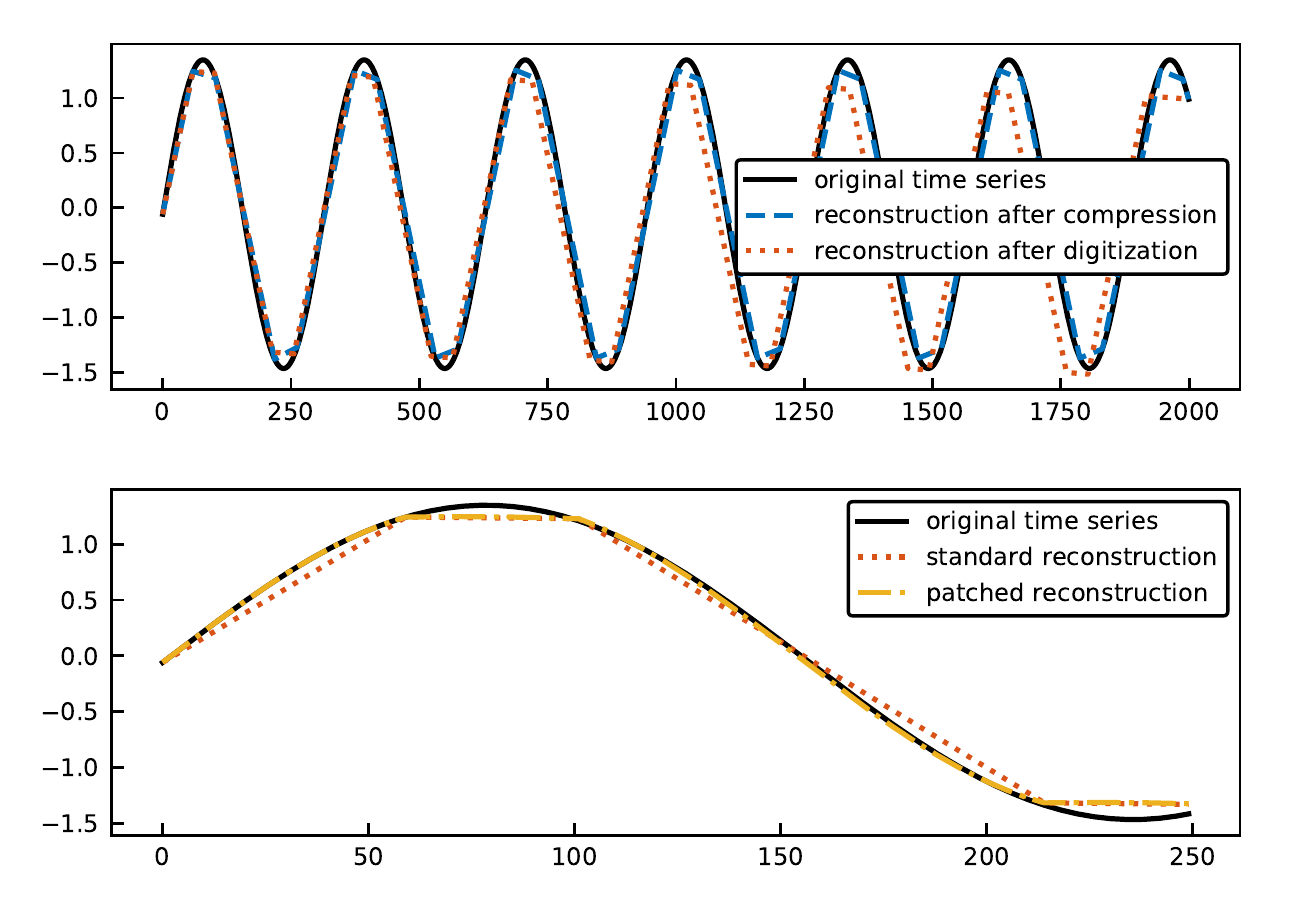}
\vspace*{-6mm}
\caption{ABBA representation of a normalized sine wave. ABBA reduces the time series of length $N=2000$ down to the sequence \texttt{dabacabacabacabacabacabacaa} of length~$m=27$ using $k=4$ symbols. Note that a unique symbol is allocated to the first segment of the time series and the symbols following this ``start-up'' phase closely follow the regularity of the sine wave.}
\label{fig:patches}
\end{figure}

ABBA is a symbolic time series  representation where the symbolic length~$m$ and the number of symbols $k$ are chosen adaptively~\cite{EG19b}. The ABBA representation is computed in two stages: \emph{compression} and \emph{digitization}. The compression stage constructs an adaptive piecewise linear  approximation of the time series. The algorithm selects $m+1$ indices 
$i_0 = 1 < i_1 < \cdots < i_m = N$ such that the time series $T$ is partitioned into $m$ pieces  $P_j = [t_{i_{j-1}}, \ldots, t_{i_j}]$, $j=1,2,\ldots,m$. On each piece $P_j$, the time series is approximated by a straight line through the end point values, represented by the tuple $(\texttt{len}_j, \texttt{inc}_j) \in \mathbb{R}^2$ defined as $\texttt{len}_j = i_{j} - i_{j-1}$ and $\texttt{inc}_j = t_{i_j} - t_{i_{j-1}}$. The sequence of tuples $(\texttt{len}_1, \texttt{inc}_1), \ldots, (\texttt{len}_m, \texttt{inc}_m)$ and the first value $t_1$ represent a polygonal chain going through the points $(i_j, t_{i_j})$ for $j=0, 1, \ldots, m$. An example output of the ABBA compression algorithm applied to a z-normalised sine wave is shown in the first plot of Figure~\ref{fig:patches}. 

During the ABBA digitization stage, the tuples $(\texttt{len}_j, \texttt{inc}_j)$ are grouped into $k$ clusters using a mean-based clustering algorithm, with each cluster assigned a symbol from the alphabet $\mathbb{A}$.
Converting from the symbolic representation back to a numeric representation requires three stages: \emph{inverse-digitization}, \emph{quantization} and \emph{inverse-compression}. The inverse-digitization stage represents each symbol by the center of the corresponding cluster, resulting in a sequence of tuples. The quantization realigns the accumulated lengths of the tuples with an integer grid. Finally, the inverse-compression stage stitches the linear pieces represented by each tuple to obtain raw time series values. It is shown in~\cite{EG19b} that this back-conversion to the raw  time series values leads to reconstruction errors that form a so-called Brownian bridge, giving ABBA its name (``adaptive Brownian bridge-based aggregation''). 

As an alternative to the polygonal chain approximation used in~\cite{EG19b}, we propose to represent each cluster by the mean of time series pieces falling in that cluster. For simplicity of exposition, suppose that the digitization procedure has returned a cluster $$S_1 = \{ (\texttt{len}_1, \texttt{inc}_1), (\texttt{len}_3, \texttt{inc}_3), (\texttt{len}_7, \texttt{inc}_7) \}$$ with cluster center $(\overline{\texttt{len}_1}, \overline{\texttt{inc}_1})$. Each of the tuples in $S_1$ corresponds to a piece of the raw time series data, $[t_{i_0}, \ldots, t_{i_1}]$, $[t_{i_2}, \ldots, t_{i_3}]$ and $[t_{i_6}, \ldots, t_{i_7}]$, respectively. We  propose to extrapolate/interpolate each of these pieces to form new time series of a common average length $\mathrm{round}(\overline{\texttt{len}_1})$. The point-wise mean of the new interpolated time series provides a smooth numerical representation for that cluster, which we refer to as a ``patch.'' The reconstruction of raw numerical time series values can now be obtained by stitching these patches in accordance with the order of symbols in the ABBA string.

This new patched ABBA reconstruction provides a visually more appealing representation of the time series as averages of shapes appearing in the raw time series are being used. The second plot in Figure~\ref{fig:patches} illustrates the difference between a standard reconstruction and a patched reconstruction on a zoomed-in version of the sine wave.

\section{Recurrent neural networks} \label{sec:lstm}
Consider a sequence $x = (\mathbf{x}^{(1)}, \mathbf{x}^{(2)}, \ldots, \mathbf{x}^{(\ell)})$, where each element  $\mathbf{x}^{(i)}\in\mathbb{R}^d$ is a vector of dimension~$d$. When training a traditional neural network on that data, we would feed in all  information about this sequence in one go. See also the illustration in Figure~\ref{fig:comp_standard}. This approach would ignore any temporal dependencies present in the sequence $x$. Furthermore, the number of weights in the network would increase linearly with the sequence length~$\ell$. 

\begin{figure}[h!t]
\centering
\begin{tikzpicture}[
    font=\sf \scriptsize,
    >=LaTeX,
    ct/.style={
        line width = .75pt,
        minimum width=2cm,
        inner sep=1pt,
        },
    gt/.style={
        circle,
        black!60!green,
        draw,
        minimum width=6mm,
        minimum height=6mm,
        inner sep=1pt
        },
    Arrow/.style={
   	    thick,
        rounded corners=.5cm,
        },
     mylabel/.style={
        font=\scriptsize\sffamily
        },
    ]
\node[gt] (n1) at (-4, 0) {$\sigma, b$};

\node[ct] (x1) at (-5.5, -2) {$\mathbf{x}^{(1)}$};
\node[ct] (x2) at (-4.5, -2) {$\mathbf{x}^{(2)}$};
\node[ct] (dot) at (-3.5, -2) {$\cdots$};
\node[ct] (xl) at (-2.5, -2) {$\mathbf{x}^{(\ell)}$};

\draw[->, Arrow, black] (x1) -- (n1);
\draw[->, Arrow, black] (x2) -- (n1);
\draw[->, Arrow, black] (xl) -- (n1);

\node[ct] (w1) at (-5.3, -1.2) {$\mathbf{w}_1^T$};
\node[ct] (w2) at (-4.5, -1.2) {$\mathbf{w}_2^T$};
\node[ct] (wl) at (-2.7, -1.2) {$\mathbf{w}_{\ell}^T$};

\node[ct] (o1) at (-4, 1) {$\sigma(\mathbf{w}_1^T\mathbf{x}_1+\mathbf{w}_2^T\mathbf{x}_2 + \cdots + \mathbf{w}_{\ell}^T\mathbf{x}_{\ell} + b)$};
\draw[->, Arrow, black] (n1) -- (o1);
\end{tikzpicture}
\vspace*{-3mm}
\caption{Graphical illustration of a single non-recurrent neuron with $\ell$ inputs of dimension $d$. The weights are $\mathbf{w}_1, \ldots, \mathbf{w}_{\ell} \in \mathbb{R}^d$, $b \in \mathbb{R}$ is a bias value, and $\sigma$ is an activation function.\label{fig:comp_standard}}
\end{figure}

\begin{figure*}[!t]
\centering
\hspace{1in}
\subfloat[]{\begin{tikzpicture}[
    font=\sf \scriptsize,
    >=LaTeX,
    ct/.style={
        line width = .75pt,
        minimum width=2cm,
        inner sep=1pt,
        },
    gt/.style={
        rectangle,
        blue,
        draw,
        minimum width=6mm,
        minimum height=6mm,
        inner sep=1pt
        },
    Arrow/.style={
	    thick,
        rounded corners=.5cm,
        },
     mylabel/.style={
        font=\scriptsize\sffamily
        },
    ]
\node[gt] (n1) at (-4, 0) {$\sigma, b$};

\node[ct] (wi1) at (-3.65, -0.9) {$\mathbf{w}^T$};
\node[ct] (x1) at (-4, -1.5) {$\mathbf{x}^{(i)}$};
\node[ct] (t1) at (-4.6, 0) {$w_h$};
\node[ct] (inv) at (-4, 1.5) {$h_i$};

\draw[->, Arrow, black] (x1) -- (n1);
\draw[->, Arrow, black] (n1) -- (inv);
\draw[->, Arrow, black] (n1.north) to[out=120, in=-120,looseness=10] (n1.south);

\node[ct] (size1) at (-4, -3.5) {};
\node[ct] (size2) at (-4, 3.8) {};
\end{tikzpicture}%
\label{fig:recurrent_A}}
\subfloat[]{\begin{tikzpicture}[
    font=\sf \scriptsize,
    >=LaTeX,
    ct/.style={
        line width = .75pt,
        minimum width=2cm,
        inner sep=1pt,
        },
    gt/.style={
        rectangle,
        blue,
        draw,
        minimum width=6mm,
        minimum height=6mm,
        inner sep=1pt
        },
    Arrow/.style={
        thick,
        rounded corners=.5cm,
        },
     mylabel/.style={
        font=\scriptsize\sffamily
        },
    ]

\node[gt](nn1) at (2,-2) {$\sigma, b$};
\node[gt](nn2) at (2,-0.5) {$\sigma, b$};
\node[ct](ddots) at (2, 1) {$\vdots$};
\node[gt](nnl) at (2,2.5) {$\sigma, b$};

\node[ct] (h0) at  (2, -3) {$0$};
\node[ct] (i1) at (2.75, -3) {$\mathbf{x}^{(1)}$};
\node[ct] (i2) at (3.5, -3) {$\mathbf{x}^{(2)}$};
\node[ct] (i3) at (4.25, -3) {$\cdots$};
\node[ct] (il) at (5, -3) {$\mathbf{x}^{(\ell)}$};
\node[ct] (o2) at (2, 3.5) {$\sigma(w_h(\cdots \sigma(w_h \sigma(\mathbf{w}^T\mathbf{x}^{(1)} + b) + \mathbf{w}^T\mathbf{x}^{(2)} + b)) \cdots) + \mathbf{w}^T\mathbf{x}^{(\ell)} + b)$};

\draw[->, Arrow, black] (h0) -- (nn1);
\draw[->, Arrow, black] (i1) -- (nn1);
\draw[->, Arrow, black] (i2) -- (nn2);
\draw[->, Arrow, black] (il) -- (nnl);
\draw[->, Arrow, black] (nn1) -- (nn2);
\draw[->, Arrow, black] (nn2) -- (ddots);
\draw[->, Arrow, black] (ddots) -- (nnl);
\draw[->, Arrow, black] (nnl) -- (o2);

\node[ct] (hw1) at (1.7, -2.5) {$w_h$};
\node[ct] (hw2) at (1.7, -1.5) {$w_h$};
\node[ct] (hw3) at (1.7, 0) {$w_h$};
\node[ct] (hw4) at (1.7, 1.5) {$w_h$};

\node[ct] (iw1) at (2.7, -2.5) {$\mathbf{w}^T$};
\node[ct] (iw2) at (3, -1.6) {$\mathbf{w}^T$};
\node[ct] (iwl) at (3.5, 0.5) {$\mathbf{w}^T$};
\end{tikzpicture}
\label{fig:recurrent_B}}
\hspace{1in}
\caption{Graphical illustrations of a single recurrent unit with $\ell$ inputs of dimension $d$. The weights are $\mathbf{w} \in \mathbb{R}^d$ and  $w_h \in \mathbb{R}$, and $b \in \mathbb{R}$ is a bias value.\label{fig:recurrent}}
\end{figure*}
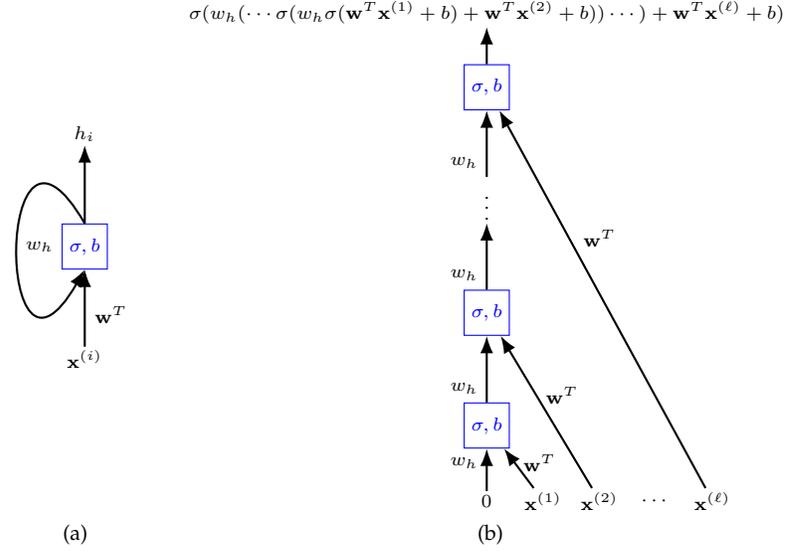

Recurrent neural networks (RNNs) are designed to process sequential data more efficiently by taking into consideration the sequential nature of the data. 
A standard recurrent neural unit can process the sequence elements one at a time, starting with first element of the sequence before feeding in the second. At each time point, the unit takes two inputs: an element of the sequence $\mathbf{x}_i \in \mathbb{R}^d$ and the output of the same unit at the previous time point, $h_{i-1} \in \mathbb{R}$.
This allows the unit to process the whole sequence with a fixed number of weights, i.e., the model is independent of the sequence length. 

A  graphical representation of a recurrent neural unit is shown in Figure~\ref{fig:recurrent_A}. If one unravels the time direction of the unit,  a graphical representation as in Figure~\ref{fig:recurrent_B} emerges, which is nothing but a traditional neural network with a specific structure and weight sharing. In these plots, the blue squares always refer to the same unit (with the same weights, bias and activation function).

Hochreiter noticed the vanishing gradient problem \cite{H98} that can occur during the weight training of a recurrent neural network. In fact, the vanishing gradient problem can occur in any deep neural network and, as shown in Figure~\ref{fig:recurrent_B}, a recurrent neural network for large $\ell$ is a very deep neural network. This led to the invention of so-called long short-term memory (LSTM) cells  \cite{HS97} and gated recurrent units (GRU) \cite{CVB14b}. 
LSTMs are popular in the machine learning community and have found many applications including handwriting recognition \cite{GLB08, GS09}, speech recognition \cite{GMH13, GJ14}, machine translation \cite{SVL14, CVB14}, and time series forecasting \cite{SWG05, CZD15, FZ16, YDJ17}.  
Many variations of the original LSTM have been proposed in \cite{GSK16}. Below   we focus on the ``literature standard LSTM'' with a forget gate and no peepholes. 
We first provide a mathematical description of a single LSTM cell in Section~\ref{sec:lstm1}, and then show how one can build a network of LSTM cells by concatenating (Section~\ref{sec:lstm2}) and composing (Section~\ref{sec:lstm3}) these cells. 

\subsection{The structure of a single LSTM cell}\label{sec:lstm1}
Let $\mathbb{U} = [0,1]$ represent the unit interval and let $\mathbb{\pm U} = [-1, 1]$. An LSTM cell has two recurrent features, denoted by $h$ and $c$, called the hidden state and the cell state, respectively. The cell, denoted by $\mathcal{L}$, is a mathematical function that takes three inputs and produces two outputs: 
\begin{equation} \label{eq:lstm_layer}
(h^{(t)}, c^{(t)}) = \mathcal{L}(h^{(t-1)}, c^{(t-1)}, \mathbf{x}^{(t)}),
\end{equation}
where $h^{(t)}, h^{(t-1)}, c^{(t)}, c^{(t-1)} \in \mathbb{\pm U}$ and $\mathbf{x}^{(t)} \in \mathbb{R}^d$. Both outputs leave the cell at time point $t$ and are fed back into that same cell at time point $t+1$. At any time point $t$, an element of the input sequence $\mathbf{x}^{(t)} \in \mathbb{R}^d$ is also fed into the cell.  

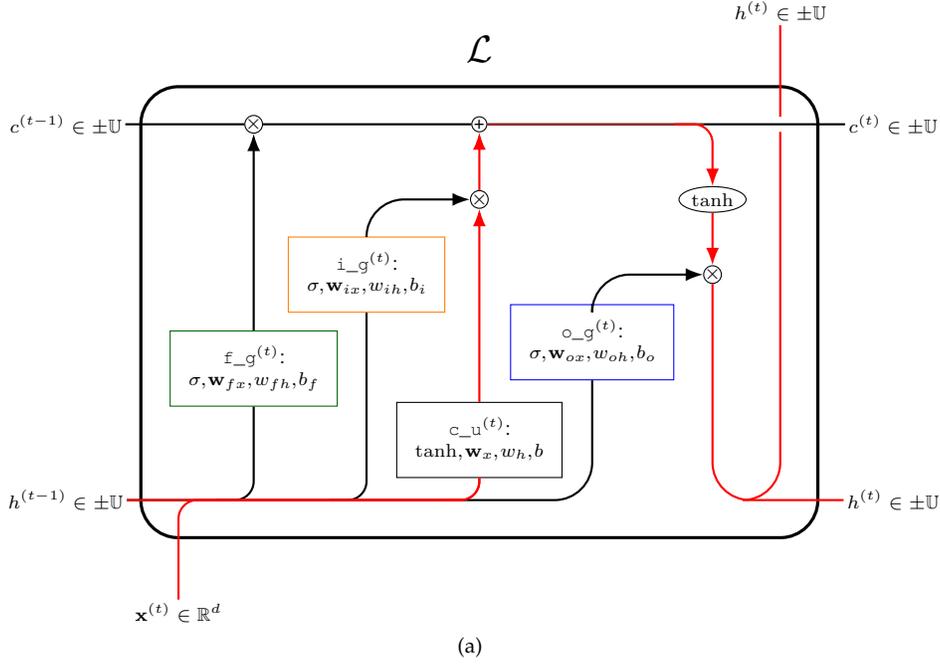
\begin{figure*}[!t]
\centering
\subfloat[]{

\newcommand*\squeezespaces[1]{
  \thickmuskip=\scalemuskip{\thickmuskip}{#1}%
  \medmuskip=\scalemuskip{\medmuskip}{#1}%
  \thinmuskip=\scalemuskip{\thinmuskip}{#1}%
  \nulldelimiterspace=#1\nulldelimiterspace
  \scriptspace=#1\scriptspace
}
\newcommand*\scalemuskip[2]{%
  \muexpr #1*\numexpr\dimexpr#2pt\relax\relax/65536\relax
} 

\begin{tikzpicture}[
    font=\sf \scriptsize,
    >=LaTeX,
    cell/.style={
        rectangle,
        rounded corners=5mm,
        draw,
        very thick,
        },
    operator/.style={
        circle,
        draw,
        inner sep=-0.5pt,
        minimum height =.2cm,
        },
    function/.style={
        ellipse,
        draw,
        inner sep=1pt
        },
    ct/.style={
        line width = .75pt,
        minimum width=1cm,
        inner sep=1pt,
        },
    gt/.style={
        rectangle,
        draw,
        minimum width=14mm,
        minimum height=10mm,
        inner sep=1pt
        },
    mylabel/.style={
        font=\scriptsize\sffamily
        },
    ArrowC1/.style={
        rounded corners=.25cm,
        thick,
        },
    ArrowC2/.style={
        rounded corners=.5cm,
        thick,
        },
    ]

    \node [cell, minimum height =6cm, minimum width=9cm] at (-0.5,0){} ;

    \node [gt, black!60!green] (ibox1) at (-3.5,-0.75) {\color{black} \begin{tabular}{c} $\texttt{f\_g}^{(t)}$: \\\mbox{$\squeezespaces{0.3}\sigma,\mathbf{w}_{fx}, w_{fh}, b_f$}\end{tabular}};
    \node [gt, orange] (ibox2) at (-2,0.5) {\color{black} \begin{tabular}{c} $\texttt{i\_g}^{(t)}$: \\\mbox{$\squeezespaces{0.3}\sigma,\mathbf{w}_{ix}, w_{ih}, b_i$}\end{tabular}};
    \node [gt, minimum width=1cm] (ibox3) at (-0.5,-1.7) {\color{black} \begin{tabular}{c} $\texttt{c\_u}^{(t)}$: \\\mbox{$\squeezespaces{0.3}\mathrm{tanh},\mathbf{w}_{x}, w_{h}, b$}\end{tabular}};
    \node [gt, blue] (ibox4) at (1,-0.4) {\color{black} \begin{tabular}{c} $\texttt{o\_g}^{(t)}$: \\\mbox{$\squeezespaces{0.3}\sigma,\mathbf{w}_{ox}, w_{oh}, b_o$}\end{tabular}};
    
   \node[ct] (f) at (-0.5, 3.5) {\Large $\mathcal{L}$};

    \node [operator] (mux1) at (-3.5,2.5) {$\times$};
    \node [operator] (add1) at (-0.5,2.5) {+};
    \node [operator] (mux2) at (-0.5,1.5) {$\times$};
    \node [operator] (mux3) at (2.6,0.5) {$\times$};
    \node [function] (func1) at (2.6,1.5) {$\mathrm{tanh}$};

    \node[ct] (c) at (-6,2.5) {$c^{(t-1)} \in \mathbb{\pm U}$};
    \node[ct] (h) at (-6,-2.5) {$h^{(t-1)} \in \mathbb{\pm U} $};
    \node[ct] (x) at (-4.5,-4) {$\mathbf{x}^{(t)} \in \mathbb{R}^d$};

    \node[ct] (c2) at (5,2.5) {$c^{(t)} \in \mathbb{\pm U}$};
    \node[ct] (h2) at (5,-2.5) {$h^{(t)} \in \mathbb{\pm U}$};
    \node[ct] (x2) at (3.5, 4) {$h^{(t)} \in \mathbb{\pm U}$};

    \node[ct] (p1) at (-0.2, -2.5) {};
    \node[ct] (p2) at (-0.9, 2.5) {};
    \node[ct] (p3) at (3, 2.5) {};

    \draw [ArrowC1] (c) -- (mux1) -- (add1) -- (c2);

    \draw [ArrowC2] (h) -| (ibox4);
    \draw [ArrowC1] (h -| ibox1)++(-0.5,0) -| (ibox1);
    \draw [ArrowC1] (h -| ibox2)++(-0.5,0) -| (ibox2);
    \draw [ArrowC2,red] (h) -- (p1);
    \draw [ArrowC1, red] (h -| ibox3)++(-0.5,0) -| (ibox3);
    \draw [ArrowC1, red] (x) -- (x |- h)-| (ibox3);

    \draw [->, ArrowC2] (ibox1) -- (mux1);
    \draw [->, ArrowC2] (ibox2) |- (mux2);
    \draw [->, ArrowC2, red] (ibox3) -- (mux2);
    \draw [->, ArrowC2] (ibox4) |- (mux3);
    \draw [->, ArrowC2, red] (mux2) -- (add1);
    \draw [->, ArrowC1, red] (add1 -| func1)++(-0.5,0) -| (func1);
    \draw [->, ArrowC2, red] (func1) -- (mux3);
    \draw [-, red] (p2) -- (p3);

    \draw (c2 -| x2) ++(0,-0.1) coordinate (i1);
    \draw [-, ArrowC2, red] (mux3) |- (h2);
    \draw [-, ArrowC2, red] (h2 -| x2)++(-0.5,0) -| (i1);
    \draw [-, ArrowC2, red] (i1)++(0,0.2) -- (x2);

\end{tikzpicture}
}%
\caption{A graphical illustration of a single LSTM cell. The red arrows show the recurrent neural unit with two $\mathrm{tanh}$ activation functions. The three gates---forget (green), input (orange), and output (blue)---control the interactions between the cell state and the hidden state. \label{fig:LSTM_cell}}
\end{figure*}

Inside the cell, the hidden state and the input vector are fed into three gates (functions), each of which produces a scalar value in $\mathbb{U}$ with the help of a sigmoid activation function: 
\begin{align*}
\texttt{f\_g}^{(t)}(\mathbf{x}^{(t)}, h^{(t-1)}) &= \sigma( \mathbf{w}_{f,x}^T \mathbf{x}^{(t)} + w_{f,h}h^{(t-1)} + b_f) \in \mathbb{U}, \\
\texttt{i\_g}^{(t)}(\mathbf{x}^{(t)}, h^{(t-1)}) &= \sigma( \mathbf{w}_{i,x}^T \mathbf{x}^{(t)} \, + w_{i,h}h^{(t-1)} + b_i) \,\,\in \mathbb{U}, \\ 
\texttt{o\_g}^{(t)}(\mathbf{x}^{(t)}, h^{(t-1)}) &= \sigma( \mathbf{w}_{o,x}^T \mathbf{x}^{(t)}\, + w_{o,h}h^{(t-1)} + b_o) \in \mathbb{U}, 
\end{align*}
\noindent where $\mathbf{w}_{f,x}, \mathbf{w}_{i,x}, \mathbf{w}_{o,x} \in \mathbb{R}^d$ and $w_{f,h}, w_{i,h}, w_{o,h}, b_f, b_i, b_o \allowbreak \in \mathbb{R}$ are weight parameters (also called weight vectors and biases, respectively). These are the  parameters to be learned during the training of the cell. 
The three gates can be interpreted as switches when their output values are near 1~(on) or 0~(off).
Another scalar function, the so-called cell update (\texttt{c\_u}), is constructed as a single neuron with a  $\tanh$ activation function 
\begin{equation*}
\texttt{c\_u}^{(t)}(\mathbf{x}^{(t)}, h^{(t-1)}) = \tanh(\mathbf{w}_x^T \mathbf{x}^{(t)} + w_h h^{(t-1)} + b) \in \mathbb{\pm U},
\end{equation*}
where $\mathbf{w}_x \in \mathbb{R}^d$ and $w_h, b \in \mathbb{R}$ are further weight parameters to be learned. The forget gate (\texttt{f\_g}) controls how much of the current cell state we should forget, the input gate (\texttt{i\_g}) controls how much of the cell update is added to the cell state, and the output gate (\texttt{o\_g}) controls how much of the modified cell state should leave the cell and become the next hidden state. Written in terms of mathematical functions, the new cell  and hidden states at time $t$ are  
\begin{align*}
c^{(t)} &= \texttt{f\_g}^{(t)} \cdot c^{(t-1)} + \texttt{i\_g}^{(t)} \cdot \texttt{c\_u}^{(t)} \in \mathbb{\pm U},\\
h^{(t)} &= \texttt{o\_g}^{(t)} \cdot \mathrm{tanh}(c^{(t)}) \in \mathbb{\pm U},
\end{align*}
where the arguments $(\mathbf{x}^{(t)}, h^{(t-1)})$ have been omitted for readability. All of these functions and parameters are encapsulated in the function $\mathcal{L}$ from Equation~\eqref{eq:lstm_layer}, and a graphical illustration of that function is given in Figure~\ref{fig:LSTM_cell}.

\begin{figure}[h!]
\centering
\begin{tikzpicture}[
    font=\sf \scriptsize,
    >=LaTeX,
    ct/.style={
        line width = .75pt,
        minimum width=1cm,
        inner sep=1pt,
        },
    gt/.style={
        rectangle,
        draw,
        minimum width=6mm,
        minimum height=4mm,
        inner sep=1pt
        },
    Arrow/.style={
        rounded corners=.5cm,
        thick,
        },
     mylabel/.style={
        font=\scriptsize\sffamily
        },
    ]
 
\node[gt] (l1) at (-1.5, 0) {$\mathcal{L}$};
\node[gt] (l2) at (0, 0) {$\mathcal{L}$};
\node[gt] (l3) at (1.5, 0) {$\mathcal{L}$};

\node[ct] (i1) at (-1.5, -1) {$\mathbf{x}^{(t)}$};
\node[ct] (i2) at (0, -1) {$\mathbf{x}^{(t+1)}$};
\node[ct] (i3) at (1.5, -1) {$\mathbf{x}^{(t+2)}$};

\node[ct] (o1) at (-1.5, 1) {};
\node[ct] (o2) at (0, 1) {};
\node[ct] (o3) at (1.5, 1) {};

\draw[->, Arrow, blue] (l1) -- (o1);
\draw[->, Arrow, blue] (l2) -- (o2);
\draw[->, Arrow, blue] (l3) -- (o3);

\draw[->, Arrow] (i1) -- (l1);
\draw[->, Arrow] (i2) -- (l2);
\draw[->, Arrow] (i3) -- (l3);

\node[ct] (z1) at (-3, 0.2) {$c^{(t-1)}$};
\node[ct] (z2) at (-3, -0.2) {$h^{(t-1)}$};
\draw [->,Arrow, black!60!green] (z1.east) to [bend left] (l1.north west);
\draw [->,Arrow, blue] (z2.east) to [bend right] (l1.south west);
\draw [->,Arrow, black!60!green] (l1.north east) to [bend left] (l2.north west);
\draw [->,Arrow, blue] (l1.south east) to [bend right] (l2.south west);
\draw [->,Arrow, black!60!green] (l2.north east) to [bend left] (l3.north west);
\draw [->,Arrow, blue] (l2.south east) to [bend right] (l3.south west);
\node[ct] (z3) at (3, 0.2) {};
\node[ct] (z4) at (3, -0.2) {};
\draw [->,Arrow, black!60!green] (l3.north east) to [bend left] (z3.west);
\draw [->,Arrow, blue] (l3.south east) to [bend right] (z4.west);
\end{tikzpicture}
\caption{Demonstrating the flexibility of an LSTM cell. The function $\mathcal L$ is capable of working with input and output sequences of any length.}
\label{fig:LSTM}
\end{figure}
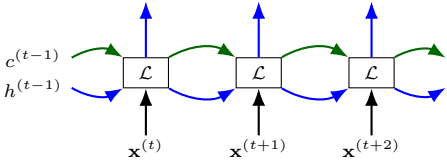

By its design using hidden states that pass through time, recurrent neural networks have the capability to take an input sequence of any length and produce an output sequence of any length; see also the graphical representation in Figure~\ref{fig:LSTM}. The user can decide at what time points to feed in the input sequence and at what time points to extract the outputs.

\subsection{A layer of LSTM cells}\label{sec:lstm2}
A layer of $n$ LSTM cells, which we denote by $\bm{\mathcal{L}}_n$, corresponds to the concatenation of $n$ cells $\mathcal{L}_1, \mathcal{L}_2, \ldots, \mathcal{L}_n$, each with a different set of internal weight parameters. That is,  
\begin{align*}
(h^{(t)}_1, c^{(t)}_1) &= \mathcal{L}_1(h^{(t-1)}_1, c^{(t-1)}_1, \mathbf{x}^{(t)}), \\
(h^{(t)}_2, c^{(t)}_2) &= \mathcal{L}_2(h^{(t-1)}_2, c^{(t-1)}_2, \mathbf{x}^{(t)}), \\
&\ \, \vdots \\
(h^{(t)}_n, c^{(t)}_n) &= \mathcal{L}_n(h^{(t-1)}_n, c^{(t-1)}_n, \mathbf{x}^{(t)}), 
\end{align*}
which can equivalently be written as
\begin{equation*}
(\mathbf{h}^{(t)}, \mathbf{c}^{(t)}) = \bm{\mathcal{L}}_n(\mathbf{h}^{(t)}, \mathbf{c}^{(t)}, \mathbf{x}^{(t)}),
\end{equation*}
where $\mathbf{h}^{(t)}, \mathbf{h}^{(t-1)}, \mathbf{c}^{(t)}, \mathbf{c}^{(t-1)} \in \mathbb{\pm U}^n$ and $\mathbf{x}^{(t)} \in \mathbb{R}^d$. The individual weight vectors and biases from each of the LSTMs can be stacked into matrices. The dot products become matrix-vector products and the scalar multiplications become element-wise multiplications. The activation functions are applied element-wise, allowing the simultaneous evaluation of a whole layer of LSTM cells. The three gates and the cell update function now contain weight matrices $W_{fx}, W_{ix}, W_{ox}, W_x \in \mathbb{R}^{n \times d}$ of size compatible  with the input vector $\mathbf{x}^{(t)} \in \mathbb{R}^d$. The stacked hidden state is of dimension $n$ and so the gates contain compatible weight matrices $W_{fh}, W_{ih}, W_{oh}, W_h \in \mathbb{R}^{n \times n}$ and bias vectors $\mathbf{b}_f, \mathbf{b}_i, \mathbf{b}_o, \mathbf{b} \allowbreak\in \mathbb{R}^n$.

Some LSTM implementations, such as  those based on the NVIDIA CUDA Deep Neural Network library for GPU processing (cudNN), use two separate bias vectors for the input and recurrent data. This takes advantage of routines that can perform fast matrix-vector products plus vector operations. In Keras the gates use a hard sigmoid activation function by default in order to behave more similarly to on-off switches~\cite{KERAS}.

\subsection{A multi-layer LSTM network}\label{sec:lstm3}
\begin{figure}[h!t]
\centering
\begin{tikzpicture}[
    font=\sf \scriptsize,
    >=LaTeX,
    ct/.style={
        line width = .75pt,
        minimum width=1cm,
        inner sep=1pt,
        },
    gt/.style={
        rectangle,
        rounded corners=0.5mm,
        draw,
        minimum width=10mm,
        minimum height=6mm,
        inner sep=1pt
        },
    Arrow/.style={
        rounded corners=.5cm,
        thick,
        },
     mylabel/.style={
        font=\scriptsize\sffamily
        },
    ]
 
\node[gt] (l3) at (0, 1.5) {$\bm{\mathcal{L}}_{n_3}$};
\node[gt] (l2) at (0, 0) {$\bm{\mathcal{L}}_{n_2}$};
\node[gt] (l1) at (0, -1.5) {$\bm{\mathcal{L}}_{n_1}$};

\node[ct] (i) at (0, -2.5) {$\mathbf{x}^{(t)}$};
\draw[->, Arrow] (i) -- (l1);

\node[ct] (z1) at (-1.5, -1.3) {$\mathbf{c}^{(t-1)}_{n_1}$};
\node[ct] (z2) at (-1.5, -1.8) {$\mathbf{h}^{(t-1)}_{n_1}$};
\draw [->,Arrow] (z1) -- (l1);
\draw [->,Arrow] (z2) -- (l1);

\node[ct] (z3) at (-1.5, 0.2) {$\mathbf{c}^{(t-1)}_{n_2}$};
\node[ct] (z4) at (-1.5, -0.2) {$\mathbf{h}^{(t-1)}_{n_2}$};
\draw [->,Arrow] (z3) -- (l2);
\draw [->,Arrow] (z4) -- (l2);

\node[ct] (z5) at (-1.5, 1.8) {$\mathbf{c}^{(t-1)}_{n_3}$};
\node[ct] (z6) at (-1.5, 1.3) {$\mathbf{h}^{(t-1)}_{n_3}$};
\draw [->,Arrow] (z5) -- (l3);
\draw [->,Arrow] (z6) -- (l3);

\node[ct] (o1) at (1.5, -1.3) {$\mathbf{c}^{(t)}_{n_1}$};
\node[ct] (o2) at (1.5, -1.8) {$\mathbf{h}^{(t)}_{n_1}$};
\draw [->,Arrow] (l1) -- (o1);
\draw [->,Arrow] (l1) -- (o2);

\node[ct] (o3) at (1.5, 0.2) {$\mathbf{c}^{(t)}_{n_2}$};
\node[ct] (o4) at (1.5, -0.2) {$\mathbf{h}^{(t)}_{n_2}$};
\draw [->,Arrow] (l2) -- (o3);
\draw [->,Arrow] (l2) -- (o4);

\node[ct] (o5) at (1.5, 1.8) {$\mathbf{c}^{(t)}_{n_3}$};
\node[ct] (o6) at (1.5, 1.3) {$\mathbf{h}^{(t)}_{n_3}$};
\draw [->,Arrow] (l3) -- (o5);
\draw [->,Arrow] (l3) -- (o6);

\node[ct] (z) at (0, 2.5) {};
\draw [->, Arrow] (l1) -- (l2);
\draw [->, Arrow] (l2) -- (l3);
\draw [->, Arrow] (l3) -- (z);
\node[ct] (l1) at (0.4, 2.2) {$\mathbf{h}^{(t)}_{n_3}$};
\node[ct] (l2) at (0.4, 0.7) {$\mathbf{h}^{(t)}_{n_2}$};
\node[ct] (l3) at (0.4, -0.8) {$\mathbf{h}^{(t)}_{n_1}$};
\end{tikzpicture}
\vspace*{-2mm}
\caption{A graphical illustration of how the states pass through a multi-layer LSTM.}
\label{fig:LSTM_stacked}
\end{figure}
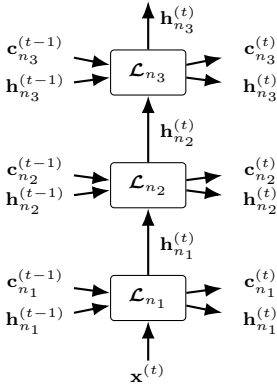

So far we have only considered a single layer of $n$~LSTM cells, called $\bm{\mathcal{L}}_n$. In practice, one often stacks multiple layers to increase the complexity of the function represented by the network. At each time point $t$, the function $\bm{\mathcal{L}}_n$ has two outputs $\mathbf{h}^{(t)}, \mathbf{c}^{(t)} \in \mathbb{\pm U}^n$. The hidden states $\mathbf{h}^{(t)}$ can be fed, sequentially, into the next layer, as shown in Figure~\ref{fig:LSTM_stacked}.

A multi-layer LSTM network can be thought of a function $\bm{\mathcal{S}}$, where 
\begin{equation*}
\begin{aligned}
&\ \, \vdots \\
(\mathbf{h}_{n_3}^{(t)}, \mathbf{c}_{n_3}^{(t)}) &= \bm{\mathcal{L}}_{n_3}(\mathbf{h}_{n_3}^{(t-1)}, \mathbf{c}_{n_3}^{(t-1)}, \mathbf{h}_{n_2}^{(t)}) \\
(\mathbf{h}_{n_2}^{(t)}, \mathbf{c}_{n_2}^{(t)}) &= \bm{\mathcal{L}}_{n_2}(\mathbf{h}_{n_2}^{(t-1)}, \mathbf{c}_{n_2}^{(t-1)}, \mathbf{h}_{n_1}^{(t)}) \\
(\mathbf{h}_{n_1}^{(t)}, \mathbf{c}_{n_1}^{(t)}) &= \bm{\mathcal{L}}_{n_1}(\mathbf{h}_{n_1}^{(t-1)}, \mathbf{c}_{n_1}^{(t-1)}, \mathbf{x}^{(t)}) \\
\end{aligned},
\end{equation*}
can be represented by 
\begin{equation*}
(\mathbf{H}^{(t)}, \mathbf{C}^{(t)}) = \bm{\mathcal{S}}(\mathbf{H}^{(t-1)}, \mathbf{C}^{(t-1)}, \mathbf{x}^{(t)}).
\end{equation*}

Each layer in the network can have a different number of cells $n_1, n_2, n_3, \ldots$, and so the hidden state  and cell state vectors may be of different dimensions. The variables $\mathbf{H}^{(t)}$ and $\mathbf{C}^{(t)}$ represent the collection of all hidden states and cell states, respectively, at time point $t$. 

We remark that in many applications, a network of stacked LSTM cells might just be a building block for a much larger model. For example, in time series forecasting, an additional final layer is used to map the output from $\mathbb{\pm U}^n$, where $n$ is the number of cells in the top layer, to time series values in $\mathbb{R}$.

\section{Training and forecasting with time series data} \label{sec:tsp}
Tuning the weights of a neural network requires a set of input/output training pairs. The inputs are feed into the network and the error between the expected output and received output is quantified via a loss function. The error is then backpropagated through the network, updating the weights via some gradient descent type scheme; see, e.g., \cite{HH19} for an introduction.

RNNs were initially proposed for language models where the length~$\ell$ and the dimension~$d$  of the sequence of inputs and outputs is pre-determined (e.g., when training on subsequences of $\ell=5$ consecutive characters of English text with $d=26$ letters). By contrast, for time series forecasting, the training set is constructed from a single time series $T = [t_1, t_2, \ldots, t_N]$ and there are no canonical lengths of the input and output sequences. Below we explain the various possibilities for feeding in sequential data into a recurrent neural network (Section~\ref{sec:tsp1}), the difference between `stateful' and `stateless' training (Section~\ref{sec:tsp2}), and finally in Section~\ref{sec:tsp3} how to produce time series forecasts.

\subsection{Feeding in sequential data}\label{sec:tsp1}
\begin{figure}[h!]
\centering
\scalebox{1}{\begin{tikzpicture}[
    font=\sf \scriptsize,
    >=LaTeX,
    ct/.style={
        line width = .75pt,
        minimum width=0.1cm,
        inner sep=1pt,
        },
    gt/.style={
        rectangle,
        draw,
        minimum width=4mm,
        minimum height=3mm,
        inner sep=1pt
        },
    Arrow/.style={
        rounded corners=.5cm,
        thick,
        },
     mylabel/.style={
        font=\scriptsize\sffamily
        },
    ]

\node[ct] (m) at (0,3) {\large $\mathcal{M}$};    
    
\draw[rounded corners] (-0.5, 1.5) rectangle (2.5, 2.5) {};
\node[gt] (l3) at (0, 2) {$\bm{\mathcal{S}}$};
\node[gt] (l4) at (1, 2) {$\bm{\mathcal{S}}$};
\node[gt] (l5) at (2, 2) {$\bm{\mathcal{S}}$};
\node[ct] (i3) at (-0, 1) {$\mathbf{x}^{(1)}$};
\node[ct] (i4) at (1, 1) {$\mathbf{x}^{(2)}$};
\node[ct] (i5) at (2, 1) {$\mathbf{x}^{(3)}$};
\node[ct] (o1) at (2, 3) {loss};

\draw[->, Arrow] (i3) -- (l3);
\draw[->, Arrow] (i4) -- (l4);
\draw[->, Arrow] (i5) -- (l5);
\draw [->,Arrow, black!60!green] (l3.north east) to [bend left] (l4.north west);
\draw [->,Arrow, blue] (l3.south east) to [bend right] (l4.south west);
\draw [->,Arrow, black!60!green] (l4.north east) to [bend left] (l5.north west);
\draw [->,Arrow, blue] (l4.south east) to [bend right] (l5.south west);
\draw [->,Arrow, blue] (l5) -- (o1);

\draw[->, dashed, Arrow, red](o1.south) -- (l5.north);
\draw[->, dashed, Arrow, red](l5.west) -- (l4.east);
\draw[->, dashed, Arrow, red](l4.west) -- (l3.east);

\node[ct] (w1) at (3.9, 2.5) {};
\node[ct] (w2) at (4.3, 2.5) {};
\node[ct] (w3) at (3.9, 2) {};
\node[ct] (w4) at (4.3, 2) {};
\node[ct] (w5) at (3.9, 1.5) {};
\node[ct] (w6) at (4.3, 1.5) {};
\draw[-, blue] (w1) -- (w2);
\draw[-, black!60!green] (w3) -- (w4);
\draw[-, dashed, red] (w5) -- (w6);
\node[ct] (t1) at (5.25, 2.5) {hidden state};
\node[ct] (t2) at (5.1, 2) {cell state};
\node[ct] (t3) at (5.85, 1.5) {backprop through time};

\node[ct] (z1) at (-1, 2.2) {$\mathbf{C}^{(0)}$};
\node[ct] (z2) at (-1, 1.8) {$\mathbf{H}^{(0)}$};
\draw [->,Arrow, black!60!green] (z1.east) to [bend left] (l3.north west);
\draw [->,Arrow, blue] (z2.east) to [bend right] (l3.south west);

\node[ct] (z3) at (3, 2.2) {$\mathbf{C}^{(5)}$};
\node[ct] (z4) at (3, 1.8) {$\mathbf{H}^{(5)}$};
\draw [->,Arrow, black!60!green] (l5.north east) to [bend left] (z3.west);
\draw [->,Arrow, blue] (l5.south east) to [bend right] (z4.west);
\end{tikzpicture}
}
\caption{An illustration of the training procedure for an LSTM network with sequences of length $\ell = 3$. The function $\mathcal{M}$ takes in the initial states and the $\texttt{input} = (\mathbf{x}^{(1)}, \mathbf{x}^{(2)}, \mathbf{x}^{(3)})$. The first two outputs of the stacked LSTM are ignored. The blue lines show the path of the hidden state and the green lines show the path of the cell state. The red dashed lines show the route taken during the backpropagation to update the weights inside~$\mathcal{S}$. The weights inside $\mathcal{S}$ receive three additive updates.}
\label{fig:M}
\end{figure}
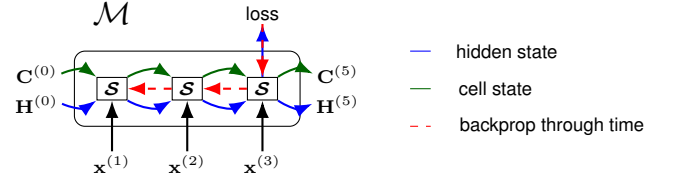

\noindent In Section~\ref{sec:lstm} we have looked at evaluating the network of LSTM cells at a single time point. 
Recall that we want to train a recurrent model on input sequences of length $\ell$, say  $\texttt{input} = (\mathbf{x}^{(1)}, \mathbf{x}^{(2)}, \ldots, \mathbf{x}^{(\ell)})$.
For simplicity of presentation, suppose that $\ell = 3$, and we want an output sequence of length one. The model is recurrent and so the function $\bm{\mathcal{S}}$ is applied three times. We can think of this procedure as a model $\mathcal{M}$ such that 
\begin{equation*}
\begin{aligned}
(\mathbf{H}^{(3)}, \mathbf{C}^{(3)}) &= \mathcal{M}(\mathbf{H}^{(0)}, \mathbf{C}^{(0)}, \texttt{input}) \\
&= \bm{\mathcal{S}}(\bm{\mathcal{S}}(\bm{\mathcal{S}}(\mathbf{H}^{(0)}, \mathbf{C}^{(0)}, \mathbf{x}^{(1)}), \mathbf{x}^{(2)}), \mathbf{x}^{(3)}).
\end{aligned}
\end{equation*} 
Figure~\ref{fig:M} illustrates how the states and input sequence are fed into the function $\mathcal{M}$ with respect to the function $\bm{\mathcal{S}}$. Note that the weights inside $\bm{\mathcal{S}}$ remain fixed when evaluating $\mathcal{M}$.

During backpropagation, each weight in $\bm{\mathcal{S}}$ receives  $\ell$ additive updates, one corresponding to each time element in the input sequence.
The length of the input sequence $\ell$, often referred to as the lag parameter, plays a critical role in defining the function $\mathcal{M}$. In time series forecasting, we want the model to have access to as many historical observations as possible. Any memory about the time series prior to the input, $\texttt{input}_i = (t_i, t_{i+1}, t_{i+2}, t_{i+3}, t_{i+4})$, must come from the cell state $\textbf{C}^{(0)}$ and the hidden state $\textbf{H}^{(0)}$. This leads to two variations of training a model containing LSTM cells known as `stateful' and `stateless' training.

\subsection{`stateful' vs `stateless' training}\label{sec:tsp2}
Suppose that the lag parameter $\ell$ has been fixed, and recall that we need to construct a training set of input/output pairs from a given time series $T = [t_1, t_2, \ldots, t_N]$. Also recall that  our model function $\mathcal{M}$ has three input arguments, $\mathbf{H}^{(0)}, \mathbf{C}^{(0)}$,  and the $\texttt{input}$ data. The training set will be constructed by an overlapping sliding window of width $\ell + 1$, giving a total of $N - \ell$ windows. The first $\ell$ values in each window form our input sequence and the trailing values form our output sequence. For example, suppose $N = 8$ and $\ell = 3$, then our training set given as 
\begin{multline*}
\big\{  [t_1, t_2, t_3\, |\, t_4],\ [t_2, t_3, t_4\, |\, t_5],\  [t_3, t_4, t_5\, |\, t_6],\ \\ [t_4, t_5, t_6\, |\, t_7],\  [t_5, t_6, t_7\, |\, t_8]  \big\},
\end{multline*}
where the vertical line partitions the inputs and the output, i.e., $[\texttt{input}_i\, |\, \texttt{output}_i]$. 

\begin{figure*}[!t]
\centering
\subfloat[]{\begin{tikzpicture}[
    font=\sf \scriptsize,
    >=LaTeX,
    ct/.style={
        line width = .75pt,
        minimum width=0.1cm,
        inner sep=1pt,
        },
    gt/.style={
        rectangle,
        draw,
        minimum width=4mm,
        minimum height=3mm,
        inner sep=1pt
        },
    Arrow/.style={
        rounded corners=.5cm,
        thick,
        },
     mylabel/.style={
        font=\scriptsize\sffamily
        },
    ]

\node[ct] (m) at (0,3) {\large $\mathcal{M}$};       
   
\draw[rounded corners] (-2.5, 1.5) rectangle (2.5, 2.5) {};
\node[gt] (l1) at (-2, 2) {$\bm{\mathcal{S}}$};
\node[gt] (l2) at (-1, 2) {$\bm{\mathcal{S}}$};
\node[gt] (l3) at (0, 2) {$\bm{\mathcal{S}}$};
\node[gt] (l4) at (1, 2) {$\bm{\mathcal{S}}$};
\node[gt] (l5) at (2, 2) {$\bm{\mathcal{S}}$};
\node[ct] (i1) at (-2, 1.2) {$t_{i}$};
\node[ct] (i2) at (-1, 1.2) {$t_{i+1}$};
\node[ct] (i3) at (-0, 1.2) {$t_{i+2}$};
\node[ct] (i4) at (1, 1.2) {$t_{i+3}$};
\node[ct] (i5) at (2, 1.2) {$t_{i+4}$};
\node[ct] (o1) at (2, 2.8) {loss};
\draw[->, Arrow] (i1) -- (l1);
\draw[->, Arrow] (i2) -- (l2);
\draw[->, Arrow] (i3) -- (l3);
\draw[->, Arrow] (i4) -- (l4);
\draw[->, Arrow] (i5) -- (l5);
\draw [->,Arrow, black!60!green] (l1.north east) to [bend left] (l2.north west);
\draw [->,Arrow, blue] (l1.south east) to [bend right] (l2.south west);
\draw [->,Arrow, black!60!green] (l2.north east) to [bend left] (l3.north west);
\draw [->,Arrow, blue] (l2.south east) to [bend right] (l3.south west);
\draw [->,Arrow, black!60!green] (l3.north east) to [bend left] (l4.north west);
\draw [->,Arrow, blue] (l3.south east) to [bend right] (l4.south west);
\draw [->,Arrow, black!60!green] (l4.north east) to [bend left] (l5.north west);
\draw [->,Arrow, blue] (l4.south east) to [bend right] (l5.south west);
\draw [->,Arrow, blue] (l5) -- (o1);

\draw[->, dashed, Arrow, red](o1.south) -- (l5.north);
\draw[->, dashed, Arrow, red](l5.west) -- (l4.east);
\draw[->, dashed, Arrow, red](l4.west) -- (l3.east);
\draw[->, dashed, Arrow, red](l3.west) -- (l2.east);
\draw[->, dashed, Arrow, red](l2.west) -- (l1.east);

\node[ct] (m) at (0,0) {\large $\mathcal{M}$};  
 
\draw[rounded corners] (-2.5, -1.5) rectangle (2.5, -0.5) {};
\node[gt] (ll1) at (-2, -1) {$\bm{\mathcal{S}}$};
\node[gt] (ll2) at (-1, -1) {$\bm{\mathcal{S}}$};
\node[gt] (ll3) at (0, -1) {$\bm{\mathcal{S}}$};
\node[gt] (ll4) at (1, -1) {$\bm{\mathcal{S}}$};
\node[gt] (ll5) at (2, -1) {$\bm{\mathcal{S}}$};
\node[ct] (ii1) at (-2, -1.8) {$t_{i+5}$};
\node[ct] (ii2) at (-1, -1.8) {$t_{i+6}$};
\node[ct] (ii3) at (-0, -1.8) {$t_{i+7}$};
\node[ct] (ii4) at (1, -1.8) {$t_{i+8}$};
\node[ct] (ii5) at (2, -1.8) {$t_{i+9}$};
\node[ct] (oo1) at (2, -0.2) {loss};
\draw[->, Arrow] (ii1) -- (ll1);
\draw[->, Arrow] (ii2) -- (ll2);
\draw[->, Arrow] (ii3) -- (ll3);
\draw[->, Arrow] (ii4) -- (ll4);
\draw[->, Arrow] (ii5) -- (ll5);
\draw [->,Arrow, blue] (ll1.north east) to [bend left] (ll2.north west);
\draw [->,Arrow, black!60!green] (ll1.south east) to [bend right] (ll2.south west);
\draw [->,Arrow, blue] (ll2.north east) to [bend left] (ll3.north west);
\draw [->,Arrow, black!60!green] (ll2.south east) to [bend right] (ll3.south west);
\draw [->,Arrow, blue] (ll3.north east) to [bend left] (ll4.north west);
\draw [->,Arrow, black!60!green] (ll3.south east) to [bend right] (ll4.south west);
\draw [->,Arrow, blue] (ll4.north east) to [bend left] (ll5.north west);
\draw [->,Arrow, black!60!green] (ll4.south east) to [bend right] (ll5.south west);
\draw [->,Arrow, blue] (ll5) -- (oo1);

\node[ct] (w1) at (2.9, 2.5) {};
\node[ct] (w2) at (3.3, 2.5) {};
\node[ct] (w3) at (2.9, 2) {};
\node[ct] (w4) at (3.3, 2) {};
\node[ct] (w5) at (2.9, 1.5) {};
\node[ct] (w6) at (3.3, 1.5) {};
\draw[-, blue] (w1) -- (w2);
\draw[-, green] (w3) -- (w4);
\draw[-, dashed, red] (w5) -- (w6);
\node[ct] (t1) at (4.2, 2.5) {hidden state};
\node[ct] (t2) at (4, 2) {cell state};
\node[ct] (t3) at (4.05, 1.5) {backprop};

\draw[->, dashed, Arrow, red](oo1.south) -- (ll5.north);
\draw[->, dashed, Arrow, red](ll5.west) -- (ll4.east);
\draw[->, dashed, Arrow, red](ll4.west) -- (ll3.east);
\draw[->, dashed, Arrow, red](ll3.west) -- (ll2.east);
\draw[->, dashed, Arrow, red](ll2.west) -- (ll1.east);

\draw[->, Arrow, black!60!green](l5.east) -- (2.8,2) -- (2.8, 0.5) -- (-2.7, 0.5) -- (-2.75, -1) -- (ll1.west);

\draw[->, Arrow, blue](l5.east) -- (2.7,2) -- (2.7, 0.6) -- (-2.8, 0.6) -- (-2.85, -1) -- (ll1.west);

\draw[->, Arrow, black!60!green](ll5.east) -- (2.6,-1) -- (2.7,-2.5);
\draw[->, Arrow, blue](ll5.east) -- (2.6,-1) -- (2.6,-2.5);

\node[ct] (z1) at (-3, 2.2) {$\mathbf{0}$};
\node[ct] (z2) at (-3, 1.8) {$\mathbf{0}$};
\draw [->,Arrow, black!60!green] (z1.east) to [bend left] (l1.north west);
\draw [->,Arrow, blue] (z2.east) to [bend right] (l1.south west);
\end{tikzpicture}
\label{fig:training_stateful}}\hspace*{4mm}
\subfloat[]{\begin{tikzpicture}[
    font=\sf \scriptsize,
    >=LaTeX,
    ct/.style={
        line width = .75pt,
        minimum width=0.1cm,
        inner sep=1pt,
        },
    gt/.style={
        rectangle,
        draw,
        minimum width=4mm,
        minimum height=3mm,
        inner sep=1pt
        },
    Arrow/.style={
        rounded corners=.5cm,
        thick,
        },
     mylabel/.style={
        font=\scriptsize\sffamily
        },
    ]

\node[ct] (m) at (0,3) {\large $\mathcal{M}$};       
   
\draw[rounded corners] (-2.5, 1.5) rectangle (2.5, 2.5) {};
\node[gt] (l1) at (-2, 2) {$\bm{\mathcal{S}}$};
\node[gt] (l2) at (-1, 2) {$\bm{\mathcal{S}}$};
\node[gt] (l3) at (0, 2) {$\bm{\mathcal{S}}$};
\node[gt] (l4) at (1, 2) {$\bm{\mathcal{S}}$};
\node[gt] (l5) at (2, 2) {$\bm{\mathcal{S}}$};
\node[ct] (i1) at (-2, 1.2) {$t_{i}$};
\node[ct] (i2) at (-1, 1.2) {$t_{i+1}$};
\node[ct] (i3) at (-0, 1.2) {$t_{i+2}$};
\node[ct] (i4) at (1, 1.2) {$t_{i+3}$};
\node[ct] (i5) at (2, 1.2) {$t_{i+4}$};
\node[ct] (o1) at (2, 2.8) {loss};
\draw[->, Arrow] (i1) -- (l1);
\draw[->, Arrow] (i2) -- (l2);
\draw[->, Arrow] (i3) -- (l3);
\draw[->, Arrow] (i4) -- (l4);
\draw[->, Arrow] (i5) -- (l5);
\draw [->,Arrow, black!60!green] (l1.north east) to [bend left] (l2.north west);
\draw [->,Arrow, blue] (l1.south east) to [bend right] (l2.south west);
\draw [->,Arrow, black!60!green] (l2.north east) to [bend left] (l3.north west);
\draw [->,Arrow, blue] (l2.south east) to [bend right] (l3.south west);
\draw [->,Arrow, black!60!green] (l3.north east) to [bend left] (l4.north west);
\draw [->,Arrow, blue] (l3.south east) to [bend right] (l4.south west);
\draw [->,Arrow, black!60!green] (l4.north east) to [bend left] (l5.north west);
\draw [->,Arrow, blue] (l4.south east) to [bend right] (l5.south west);
\draw [->,Arrow, blue] (l5) -- (o1);

\draw[->, dashed, Arrow, red](o1.south) -- (l5.north);
\draw[->, dashed, Arrow, red](l5.west) -- (l4.east);
\draw[->, dashed, Arrow, red](l4.west) -- (l3.east);
\draw[->, dashed, Arrow, red](l3.west) -- (l2.east);
\draw[->, dashed, Arrow, red](l2.west) -- (l1.east);

\node[ct] (m) at (0,0) {\large $\mathcal{M}$};  
 
\draw[rounded corners] (-2.5, -1.5) rectangle (2.5, -0.5) {};
\node[gt] (ll1) at (-2, -1) {$\bm{\mathcal{S}}$};
\node[gt] (ll2) at (-1, -1) {$\bm{\mathcal{S}}$};
\node[gt] (ll3) at (0, -1) {$\bm{\mathcal{S}}$};
\node[gt] (ll4) at (1, -1) {$\bm{\mathcal{S}}$};
\node[gt] (ll5) at (2, -1) {$\bm{\mathcal{S}}$};
\node[ct] (ii1) at (-2, -1.8) {$t_{i+1}$};
\node[ct] (ii2) at (-1, -1.8) {$t_{i+2}$};
\node[ct] (ii3) at (-0, -1.8) {$t_{i+3}$};
\node[ct] (ii4) at (1, -1.8) {$t_{i+4}$};
\node[ct] (ii5) at (2, -1.8) {$t_{i+5}$};
\node[ct] (oo1) at (2, -0.2) {loss};
\draw[->, Arrow] (ii1) -- (ll1);
\draw[->, Arrow] (ii2) -- (ll2);
\draw[->, Arrow] (ii3) -- (ll3);
\draw[->, Arrow] (ii4) -- (ll4);
\draw[->, Arrow] (ii5) -- (ll5);
\draw [->,Arrow, blue] (ll1.north east) to [bend left] (ll2.north west);
\draw [->,Arrow, black!60!green] (ll1.south east) to [bend right] (ll2.south west);
\draw [->,Arrow, blue] (ll2.north east) to [bend left] (ll3.north west);
\draw [->,Arrow, black!60!green] (ll2.south east) to [bend right] (ll3.south west);
\draw [->,Arrow, blue] (ll3.north east) to [bend left] (ll4.north west);
\draw [->,Arrow, black!60!green] (ll3.south east) to [bend right] (ll4.south west);
\draw [->,Arrow, blue] (ll4.north east) to [bend left] (ll5.north west);
\draw [->,Arrow, black!60!green] (ll4.south east) to [bend right] (ll5.south west);
\draw [->,Arrow, blue] (ll5) -- (oo1);


\draw[->, dashed, Arrow, red](oo1.south) -- (ll5.north);
\draw[->, dashed, Arrow, red](ll5.west) -- (ll4.east);
\draw[->, dashed, Arrow, red](ll4.west) -- (ll3.east);
\draw[->, dashed, Arrow, red](ll3.west) -- (ll2.east);
\draw[->, dashed, Arrow, red](ll2.west) -- (ll1.east);

\node[ct] (z1) at (-3, 2.2) {$\mathbf{0}$};
\node[ct] (z2) at (-3, 1.8) {$\mathbf{0}$};
\draw [->,Arrow, black!60!green] (z1.east) to [bend left] (l1.north west);
\draw [->,Arrow, blue] (z2.east) to [bend right] (l1.south west);

\node[ct] (zz1) at (-3, -0.8) {$\mathbf{0}$};
\node[ct] (zz2) at (-3, -1.2) {$\mathbf{0}$};
\draw [->,Arrow, black!60!green] (zz1.east) to [bend left] (ll1.north west);
\draw [->,Arrow, blue] (zz2.east) to [bend right] (ll1.south west);

\node[ct] (inv1) at (2.7,-2.5) {};

\end{tikzpicture}
\label{fig:training_stateless}}
\caption{An illustration of  `stateful' (a) and `stateless' (b) training procedures of an LSTM network when trained on sequences of length~$\ell = 5$. When `stateful', the initial input states are zero vectors and the output states are passed as input for the next training step. When `stateless', the input states are always zero vectors and the output states of $\mathcal{M}$ remain unused.\label{fig:training}}
\end{figure*}
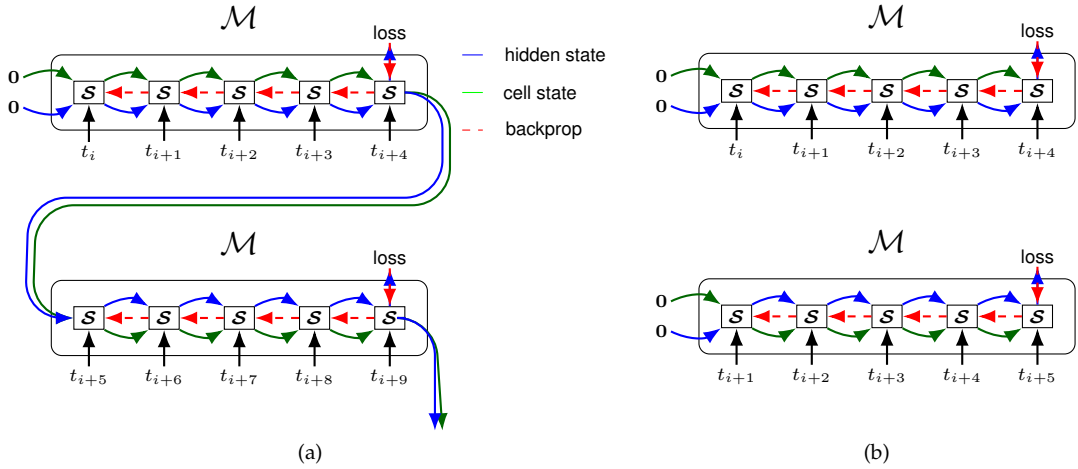

We now discuss the difference between `stateful' and `stateless' training. It might be helpful to view this together with the graphical illustration given in~Figure~\ref{fig:training}. 

A `stateless' training procedure always takes the initial states $\mathbf{H}^{(0)}$ and $\mathbf{C}^{(0)}$ to be zero. Each element of the training set is independent of all other elements, and so the training set can be shuffled after each iteration. This formulation allows mini-batch operations.

A `stateful' training procedure tries to fully exploit the memory of the network by feeding in the output states, $\mathbf{H}^{(\ell)}$ and $\mathbf{C}^{(\ell)}$, from one input/output pair as the next input state. To preserve the chronological order of the time series, the elements of the training set must be partitioned into $\ell$ groups, and the elements within each group must remain ordered. For example, when $N = 8$ and $\ell = 3$ as above, the groups are:
\begin{itemize}
\item[--] $[t_1, t_2, t_3\, |\, t_4],\  [t_4, t_5, t_6\, |\, t_7]$,
\item[--] $[t_2, t_3, t_4\, |\, t_5],\  [t_5, t_6, t_7\, |\, t_8]$, 
\item[--] $[t_3, t_4, t_5\, |\, t_6]$.
\end{itemize}
During training, the states are set to zero at the start of each group, but not as the ordered elements of each group are being fed in. The group elements are ordered such that the inputs follow the ordering of the time series data. The training of one group is independent of another, and so the groups can be shuffled between iterations. This formulation prevents mini-batch operations as the ordered groups have different cardinalities (as in our example).  

\subsection{Producing forecasts}\label{sec:tsp3}
After the recurrent model, say $\mathcal{F}$, has been trained, we would like to produce out-of-sample multi-step time series forecasts. In other words, given the time series  $T = [t_1,t_2,\ldots,t_N]$, we would like to forecast $k$ time points into the future to obtain $\hat{t}_{N+1}, \hat{t}_{N+2}, \ldots, \hat{t}_{N+k}$. There are three different ways to produce such $k$-step forecasts~\cite{AES99, MSA18}. 


\begin{description}
\item[\textbf{Iterated forecasting:}] \hfill \\ Train a `many-to-one' function $\mathcal{F}$ such that
\begin{equation*}
t_{i+\ell} \approx \mathcal{F}(t_{i}, \ldots, t_{i+\ell-1})
\end{equation*}
for $i = 1, 2, \ldots, N-\ell$. A $k$-step forecast can be made by iteratively making one-step forecasts using the previously forecasted values, i.e., 
\begin{equation*}
\begin{aligned}
\hat{t}_{N+1} &:= \mathcal{F}(t_{N-\ell+1}, \ldots, t_{N-1}, t_{N}) \\
\hat{t}_{N+2} &:= \mathcal{F}(t_{N-\ell+2}, \ldots, t_{N}, \hat t_{N+1}) \\
&\ \ \vdots \\
\hat{t}_{N+k} &:= \mathcal{F}(\hat{t}_{N+k-\ell+1}, \ldots, \hat{t}_{N+k-2}, \hat{t}_{N+k-1}).
\end{aligned}
\end{equation*}
An iterated forecast has the advantage of not requiring $k$ to be specified in advance, but it can suffer from accumulated forecast errors.\\[-2mm]

\item[\textbf{Direct forecasting:}] \hfill \\ Train a `many-to-many' function $\mathcal{F}$ such that
\begin{equation*}
(t_{i+\ell}, \ldots, t_{i+\ell+k-1})  \approx \mathcal{F}(t_{i}, \ldots, t_{i+\ell-1})
\end{equation*}
for $i = 1, 2, \ldots, N-\ell-k+1$. This requires an a-priori choice of the value~$k$. A $k$-step forecast is made as
\begin{equation*}
(\hat{t}_{N+1}, \ldots, \hat{t}_{N+k}) := \mathcal{F}(t_{N-\ell +1}, \ldots, t_{N}).
\end{equation*}

\item[\textbf{Multi-neural network forecasting:}] \hfill \\ Train $k$ `many-to-one' functions $\mathcal{F}_1, \ldots, \mathcal{F}_k$ such that
\begin{equation*}
\begin{aligned}
t_{i+\ell} &\approx \mathcal{F}_1(t_{i}, \ldots, t_{i+\ell-1}) \\
t_{i+\ell+1} &\approx \mathcal{F}_2(t_{i}, \ldots, t_{i+\ell-1}) \\
&\ \, \vdots \\
t_{i+\ell+k-1} &\approx \mathcal{F}_k(t_{i}, \ldots, t_{i+\ell-1}), \\
\end{aligned}
\end{equation*}
for $i = 1, 2, \ldots, N-\ell-k+1$. This also requires an a-priori choice of the value~$k$. A $k$-step forecast can be made by evaluating $\mathcal{F}_1, \ldots, \mathcal{F}_k$ at $(t_{N-\ell +1}, \ldots, t_{N-1}, t_{N})$.
\end{description}

\section{Raw vs symbolic forecasting} \label{sec:numvssym}
LSTMs have demonstrated their effectiveness for character-based sequence generation in a number of applications. In a typical setup, each of $k$~symbols is represented as a  vector in $\{0, 1\}^k$ by one-hot encoding. That is, each symbol corresponds to a binary vector that has only zero entries values except for an entry~1 at the index corresponding to that symbol. The sequence of binary vectors is then fed into an LSTM network, the final layer of which contains $k$ neurons with a softmax activation function. This final layer outputs a vector of probabilities that sum to one. A categorical cross-entropy loss function is used to compare the produced probabilities against that of the one-hot encoded output string. The symbol with the highest probability is used as the forecast. 

Here we propose to exploit the strengths of LSTMs for character-based sequence generation by training them on ABBA symbolic representations of time series.  The LSTM network will forecast strings which are then converted back to numerical time series values using the patching procedure described in Section~\ref{sec:sym}. We refer to this combination as ABBA-LSTM. 

There are various advantages of using a symbolic representation, such as ABBA, in combination with a machine learning model, such as an LSTM network.  Firstly, the dimensional reduction of the raw time series data to just $k$ characters allows for a  faster LSTM training, without sacrificing the prediction accuracy. Secondly, we observe that by treating the prediction task as a discrete sequencing problem instead of a regression problem,  the sensitivity of the model to the choice of parameters is reduced. Thirdly, the new ABBA patching procedure restricts the produced outputs to previously seen patches of the raw time series, producing visually more appealing forecasts. We will demonstrate these advantages in this and the following section.

\subsection{Experimental setup}
\begin{figure*}[!t]
\centering
\subfloat[]{\begin{tikzpicture}[
    font=\sf \scriptsize,
    >=LaTeX,
    ct/.style={
        line width = .75pt,
        minimum width=1cm,
        inner sep=1pt,
        },
    gt/.style={
        rectangle,
        fill=black!60!green,
        draw,
        minimum width=4mm,
        minimum height=3mm,
        inner sep=1pt
        },
    bt/.style={
        circle,
        fill=blue,
        draw,
        minimum width=4mm,
        minimum height=3mm,
        inner sep=1pt
        },
    Arrow/.style={
        rounded corners=.5cm,
        thick,
        },
     mylabel/.style={
        font=\scriptsize\sffamily
        },
    ]

\node[ct] (t1) at (0, -1.5) {$t_{i} \in \mathbb{R}$};

\node[gt] (l11) at (-2, 0) {};
\node[gt] (l12) at (-1, 0) {};
\node[ct] (l13) at (0, 0) {$\cdots$};
\node[gt] (l14) at (1, 0) {};
\node[gt] (l15) at (2, 0) {};

\node[gt] (l21) at (-2, 1.5) {};
\node[gt] (l22) at (-1, 1.5) {};
\node[ct] (l23) at (0, 1.5) {$\cdots$};
\node[gt] (l24) at (1, 1.5) {};
\node[gt] (l25) at (2, 1.5) {};

\node[bt] (n1) at (0, 3) {};

\draw[->, Arrow] (t1) -- (l11);
\draw[->, Arrow] (t1) -- (l12);
\draw[->, Arrow] (t1) -- (l14);
\draw[->, Arrow] (t1) -- (l15);

\draw[->, Arrow] (l11) -- (l21);
\draw[->, Arrow] (l11) -- (l22);
\draw[->, Arrow] (l11) -- (l24);
\draw[->, Arrow] (l11) -- (l25);
\draw[->, Arrow] (l12) -- (l21);
\draw[->, Arrow] (l12) -- (l22);
\draw[->, Arrow] (l12) -- (l24);
\draw[->, Arrow] (l12) -- (l25);
\draw[->, Arrow] (l14) -- (l21);
\draw[->, Arrow] (l14) -- (l22);
\draw[->, Arrow] (l14) -- (l24);
\draw[->, Arrow] (l14) -- (l25);
\draw[->, Arrow] (l15) -- (l21);
\draw[->, Arrow] (l15) -- (l22);
\draw[->, Arrow] (l15) -- (l24);
\draw[->, Arrow] (l15) -- (l25);

\draw[->, Arrow] (l21) -- (n1);
\draw[->, Arrow] (l22) -- (n1);
\draw[->, Arrow] (l24) -- (n1);
\draw[->, Arrow] (l25) -- (n1);

\node[ct] (space) at (0,4) {};

\draw[rounded corners] (-2.5, -0.4) rectangle (2.5, 2) {};
\node[ct] (t) at (3, 0.75) {$\bm{\mathcal{S}}$};
\end{tikzpicture}
}\hspace*{8mm}
\subfloat[]{\begin{tikzpicture}[
    font=\sf \scriptsize,
    >=LaTeX,
    ct/.style={
        line width = .75pt,
        minimum width=1cm,
        inner sep=1pt,
        },
    gt/.style={
        rectangle,
        fill=black!60!green,
        draw,
        minimum width=4mm,
        minimum height=3mm,
        inner sep=1pt
        },
    st/.style={
    	rounded rectangle,
    	draw,
    	minimum width=20mm,
    	minimum height=3mm,
		inner sep=1pt    	
    	},
    bt/.style={
        circle,
        fill=blue,
        draw,
        minimum width=4mm,
        minimum height=3mm,
        inner sep=1pt
        },
    Arrow/.style={
        rounded corners=.5cm,
        thick,
        },
     mylabel/.style={
        font=\scriptsize\sffamily
        },
    ]
    
\node[ct] (input) at (0, -3) {$t_i \in \mathbb{R}^k$};

\node[gt] (l11) at (-2, -1.5) {};
\node[gt] (l12) at (-1, -1.5) {};
\node[ct] (d1) at (0, -1.5) {$\cdots$};
\node[gt] (l13) at (1, -1.5) {};
\node[gt] (l14) at (2, -1.5) {};

\node[gt] (l21) at (-2, 0) {};
\node[gt] (l22) at (-1, 0) {};
\node[ct] (d2) at (0, 0) {$\cdots$};
\node[gt] (l23) at (1, 0) {};
\node[gt] (l24) at (2, 0) {};

\node[bt] (l31) at (-1, 1.5) {};
\node[ct] (l32) at (0, 1.5) {$\cdots$};
\node[bt] (l33) at (1, 1.5) {};

\node[st] (softmaxlayer) at (0,2.5) {softmax};

\draw[->, Arrow] (input) -- (l11);
\draw[->, Arrow] (input) -- (l12);
\draw[->, Arrow] (input) -- (l13);
\draw[->, Arrow] (input) -- (l14);

\draw[->, Arrow] (l11) -- (l21);
\draw[->, Arrow] (l11) -- (l22);
\draw[->, Arrow] (l11) -- (l23);
\draw[->, Arrow] (l11) -- (l24);

\draw[->, Arrow] (l12) -- (l21);
\draw[->, Arrow] (l12) -- (l22);
\draw[->, Arrow] (l12) -- (l23);
\draw[->, Arrow] (l12) -- (l24);

\draw[->, Arrow] (l13) -- (l21);
\draw[->, Arrow] (l13) -- (l22);
\draw[->, Arrow] (l13) -- (l23);
\draw[->, Arrow] (l13) -- (l24);

\draw[->, Arrow] (l14) -- (l21);
\draw[->, Arrow] (l14) -- (l22);
\draw[->, Arrow] (l14) -- (l23);
\draw[->, Arrow] (l14) -- (l24);

\draw[->, Arrow] (l21) -- (l31);
\draw[->, Arrow] (l22) -- (l31);
\draw[->, Arrow] (l23) -- (l31);
\draw[->, Arrow] (l24) -- (l31);

\draw[->, Arrow] (l21) -- (l33);
\draw[->, Arrow] (l22) -- (l33);
\draw[->, Arrow] (l23) -- (l33);
\draw[->, Arrow] (l24) -- (l33);

\draw[->, Arrow] (l31) -- (softmaxlayer.south west);
\draw[->, Arrow] (l33) -- (softmaxlayer.south east);

\draw[rounded corners] (-2.5, -2) rectangle (2.5, 0.5) {};
\node[ct] (t) at (3, -0.75) {$\bm{\mathcal{S}}$};
\end{tikzpicture}
}
\caption{Network configurations used in our experiments. The green rectangles represent LSTM cells and the blue circles are naked neurons without an activation function. Both models contain two layers, each with  $c$ LSTM cells. The raw LSTM model has an output layer with a single neuron and the ABBA-LSTM model has a final layer with $k$ neurons.\label{fig:setup}}
\end{figure*}
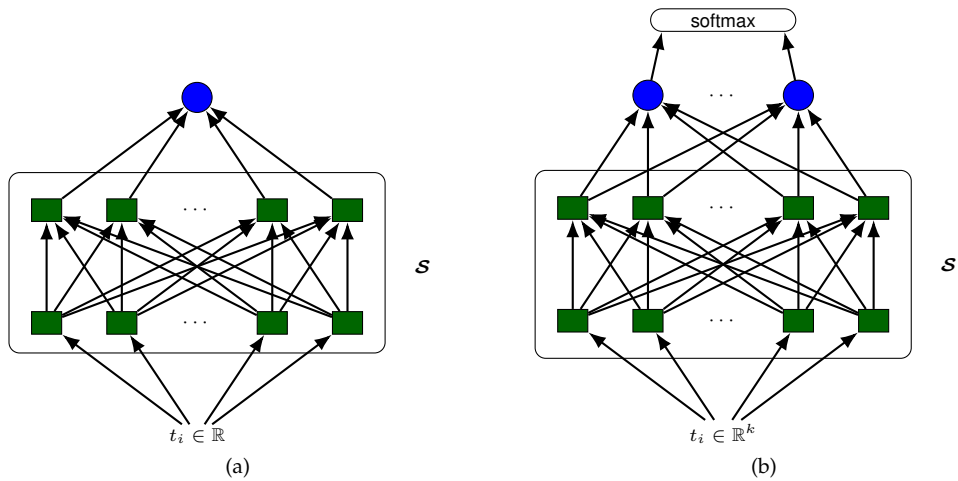

For the remainder of this paper, all LSTM models contain contain two initial layers  having $c$ cells. For the LSTM model working with the raw time series values (referred to as `raw LSTM'), an additional final layer containing a naked neuron with no activation function (or, equivalently, using $\sigma(x)=x$) maps the final hidden state $\mathbb{\pm U}^c$ to $\mathbb{R}$. This is the numerical value we consider as the time series forecast. For the symbolic ABBA-LSTM model, a final layer of $k$ neurons is added, with $k$ corresponding to the cardinality of the alphabet, followed by a softmax activation function. The model configurations are visualized in Figure~\ref{fig:setup}.

In all experiments we use the Keras default LSTM initialisation of weights, that is, the recurrent weights are initialized as random orthogonal matrices and all other weights are initialised with the Xavier uniform initializer~\cite{GB10}. The biases are initialized as zeros and the activation functions of the gates are standard sigmoids. We train using the Adam optimizer~\cite{KB14} with an early stopping criterion to control the number of iterations. We specify a patience parameter~$p$; that is, we train until there is no decrease in the loss function for $p$~consecutive iterations. After the training is completed, we backtrack the weights to the values they took when the loss was smallest. The learning rate remains fixed at $0.001$. We use the mean squared error (MSE) loss function for the raw LSTM model, and a categorical cross entropy loss function for the ABBA-LSTM model. All computations have been performed on a standard desktop machine with with 16\,GB of RAM and an Intel i7-6700 processor running at 3.4\,GHz. All experiments use Python 3.7, Keras version 2.2.4 with Tensorflow version 1.15.2 backend or PyTorch version 1.4.0.

\subsection{Study of parameter sensitivity}\label{sec:sine}
Choosing the value of the lag parameter $\ell$ is tricky. As $\ell$ increases, the size of the training set decreases. And if $\ell$ is too small, the model may struggle to learn the long-term behaviour of the time series. 
In our first experiment, we compare the raw LSTM model against the ABBA-LSTM model on the problem of forecasting from $N=1000$ samples of a sine wave with $n$ full oscillations using both a stateful and stateless training procedure. That is, our training time series values are given as $t_i = \sin(2\pi i n/N)$ for $i=1,\ldots,N$. Note that $n$ can be interpreted as the frequency of the sine wave. 
In this experiment we use a fixed lag parameter~of $\ell=50$ (time series values) for the raw LSTM model and $\ell=5$ (symbols) for the ABBA-LSTM model. 
Note that, alternatively, we could have  fixed the frequency~$n$ and vary the lag parameter $\ell$, but this would amount to changing the model rather than the training data, which would make performance comparisons meaningless.

Learning the time behaviour of a sine wave appears to be a trivial task, however, it turns out to be a difficult problem for the following reasons:
\begin{enumerate}
\item For low frequency sine waves, simply predicting $t_{i+1} = t_{i}$ already gives small values of the loss function.
\item High frequency sine waves sampled with fewer than two points per wave-length  appear as noise.
\item If the lag parameter $\ell$ is significantly smaller than the wave length, the model is trained on near linear segments.
\end{enumerate}

Both models have $c = 50$ cells per layer and are trained with a patience of $p = 50$. 
For each frequency $n=1,\ldots,100$ we train five models, each initialized using a different seed for the random number generator. (Once the seed is fixed, the remaining computations are fully deterministic.) Given a specific seed, the model's initial weights are identical regardless of the value~$n$. The stateful training procedure does not allow batch training, and so we use a batch size equal to $1$ for both the stateful and stateless training. After a model has been trained, we perform an iterated multi-step forecast to predict the next $k = 200$ time series values.

\begin{figure*}[!t]
\centering
\subfloat[]{\includegraphics[scale=1]{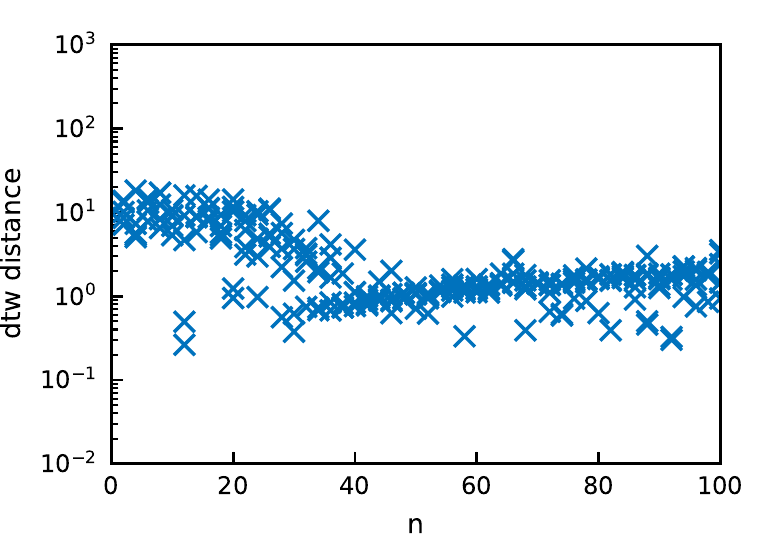}%
\label{fig:sin_stateful_raw}}
\subfloat[]{\includegraphics[scale=1]{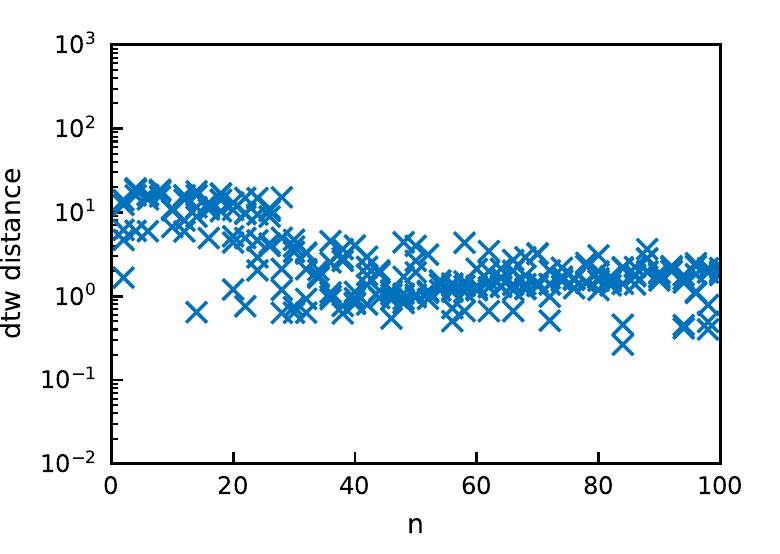}%
\label{fig:sin_stateless_raw}}

\subfloat[]{\includegraphics[scale=1]{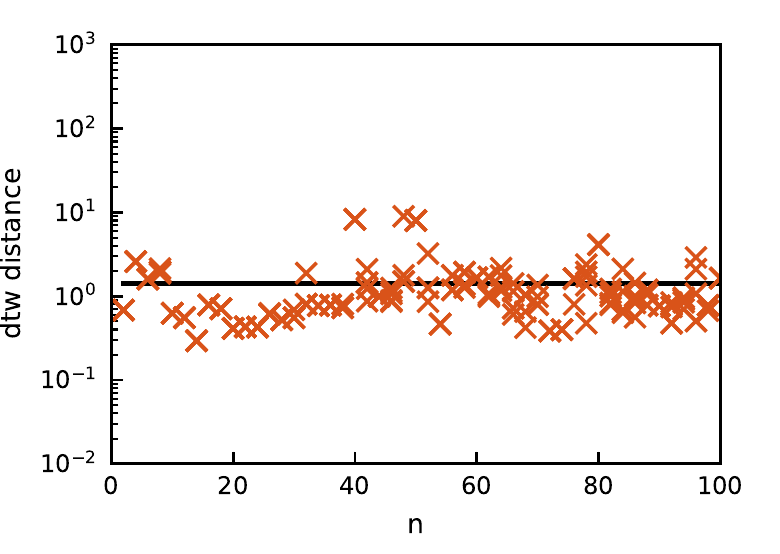}%
\label{fig:sin_stateful_ABBA}}
\subfloat[]{\includegraphics[scale=1]{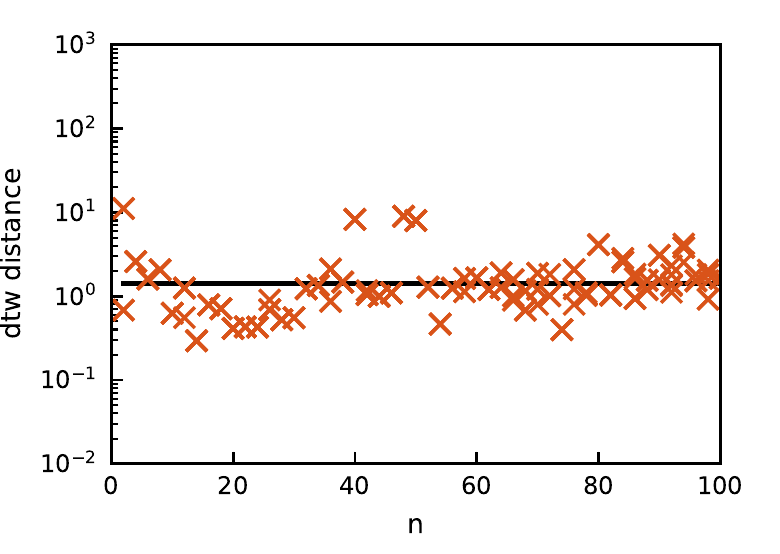}%
\label{fig:sin_stateless_ABBA}}
\caption{Experiment comparing the accuracy of raw LSTM (a, b) and ABBA-LSTM (c, d) models for forecasting sine waves of different frequency $n$ trained in stateful (a, c) and stateless (b, d) modes, respectively. The lag parameter is kept fixed in all cases and, for each $n$,  there are five models with different random initializations of the weights. The horizontal lines in (c) and (d) indicate the compression tolerance used in ABBA.\label{fig:sine}}
\end{figure*}

The results of the comparison are shown in Figure~\ref{fig:sine}, showing the dynamic time warping (DTW) distance~\cite{SC78} between the forecasts and the expected data (again a sine wave) after stateful and stateless training on the raw and symbolic data, respectively. We observe that stateless training generally results in larger DTW distances than stateful training. The raw LSTM forecasts generally have a larger DTW distance than ABBA-LSTM forecasts. The raw LSTM models are also rather sensitive to the frequency $n$, and there is a clear drop in DTW distance for a value of about $n = 40$, which corresponds to the $\ell = 50$ time series values in the sliding windows covering two full oscillations of the sine wave. 

By contrast, the performance of the stateful ABBA-LSTM model appears to be more robust with respect to changes in~$n$. This is because the ABBA string representation of a sine wave is roughly independent of the frequency~$n$. We also observe that the forecasting accuracy remains fairly close to the default tolerance of $\text{\texttt{tol}} = 0.1\sqrt{k}\approx 1.41$ used in the ABBA compression phase; see~\cite[Section~4.1]{EG19b} for details. This tolerance level is indicated by the horizontal lines in Figure~\ref{fig:sine} (c) and (d).

\subsection{Need for pre-processing}\label{sec:trend}
A raw LSTM network trained on numerical data is unlikely to forecast any values outside the numerical range of the training set. It is therefore recommended to remove (linear) time series trends before the training, as the forecasts will be poor otherwise. We  demonstrate this by considering a linearly increasing time series of length $N=200$ with values in the interval $[0, 0.5]$. We use a stateful training procedure with a lag  $\ell = 20$ and a patience $p = 10$. We repeat the training ten times, using different seeds for the random initialization of the weights. 

\begin{figure*}[!t]
\centering
\subfloat[]{\includegraphics[scale=1]{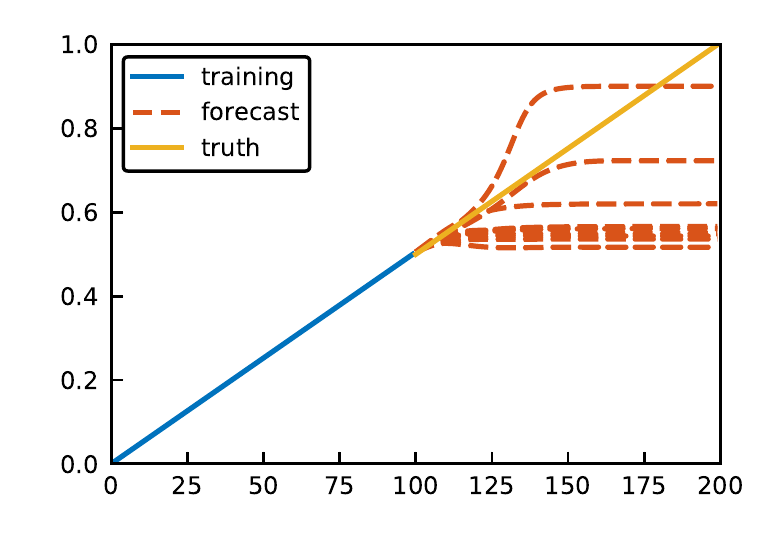}%
\label{fig:trend_numeric}}
\subfloat[]{\includegraphics[scale=1]{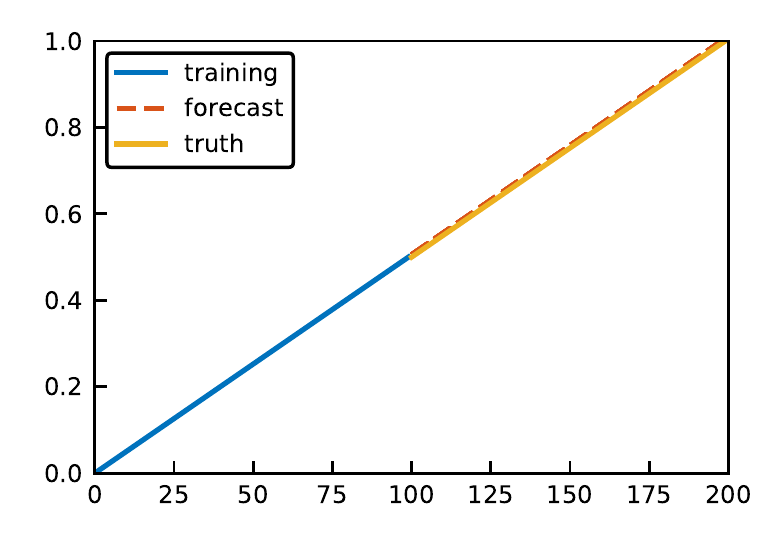}%
\label{fig:trend_symbolic}}
\caption{Demonstration of a raw LSTM model (a) struggling to forecast a linear trend, as opposed to an ABBA-LSTM model (b).}
\end{figure*}

Figure~\ref{fig:trend_numeric} shows how the LSTM model trained on the raw data fails to forecast values much greater than $0.5$. An obvious solution would be to difference the data before feeding it into the raw LSTM model, thereby removing the linear trend. However, with noisy data this can be problematic as differencing generally amplifies the noise level. The ABBA symbolic representation, on the other hand,  uses the time series increments (instead of its values) and can therefore capture linear trends directly. This allows ABBA-LSTM to forecast numerical values outside the original training range as shown in Figure~\ref{fig:trend_symbolic}.

\subsection{Shape-constraint forecasts} \label{sec:bool}
The numerical outputs of a raw LSTM model are not constrained to shapes of the original time series data and, in principle, the model can predict any value in $\mathbb{R}$. The outputs of an ABBA-LSTM model are restricted to patches of previously seen time series values. In some applications, where forecasts have to ``look natural,'' this can be beneficial. 

\begin{figure*}[!t]
\centering
\subfloat[]{\includegraphics[scale=1]{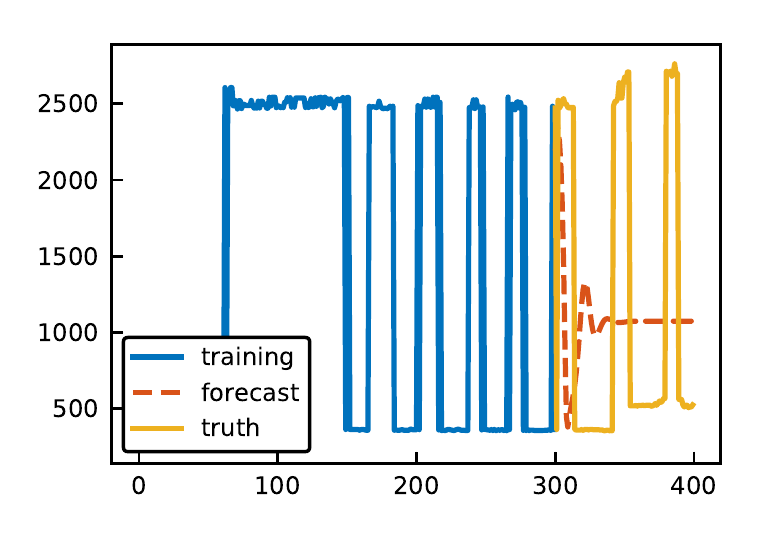}%
\label{fig:round_numeric}}
\subfloat[]{\includegraphics[scale=1]{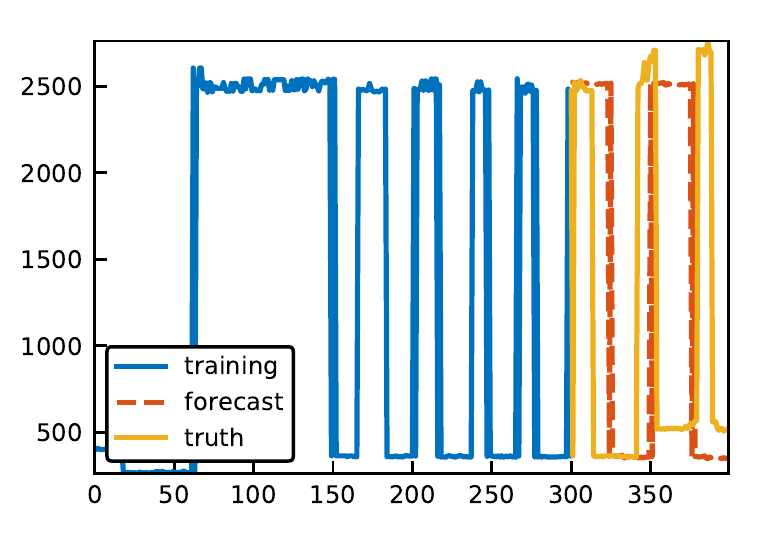}%
\label{fig:round_symbolic}}
\caption{LSTM forecast of length $200$ for the \texttt{HouseTwenty} dataset using a raw LSTM model (a) and an ABBA-LSTM model (b), respectively.\label{fig:round}}
\end{figure*}

We consider a subsequence of a time series from the \texttt{HouseTwenty} dataset in the UCR Time Series Classification archive \cite{UCRArchive}. 
As shown in Figure~\ref{fig:round}, the time series values switch between the intervals $[340, 370]$ and $[2450, 2550]$. We train LSTM models with $c=50$ cells per layer, using a lag of $\ell=50$ for the raw LSTM model and $\ell=5$ for the ABBA-LSTM model, and a patience $p = 10$ in both cases. Figure~\ref{fig:round_numeric} shows the raw LSTM forecasts, oscillating around the mean of the time series range, whereas the ABBA-LSTM forecasts, shown in Figure~\ref{fig:round_symbolic}, look more alike the original data.

\section{Performance comparison} \label{sec:rwe}
We now compare the forecasting performance of the raw LSTM and ABBA-LSTM models on time series contained in the M3 competition dataset~\cite{MSA18} and the UCR Classification Archive~\cite{UCRArchive}.

\subsection{M3 competition}
A commonly used forecasting dataset is the M3 competition data set~\cite{MSA18}, which contains $1428$ time series of lengths $N$ between $68$ and $144$. As a general observation we note that these time series are very short and it is questionable whether they constitute a reasonable training set for any machine learning-type model. This problem is even amplified when using ABBA's dimensional reduction which results in very short symbolic strings.

\begin{figure}[h!]
\centering
\vspace*{3mm}
\includegraphics[scale=1]{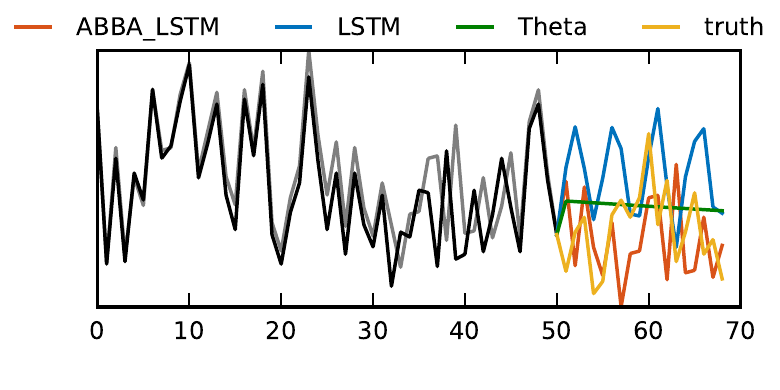}
\vspace*{-4mm}
\caption{Forecasts of length $18$ of the time series \texttt{N1500} from the M3 data set using various models. The grey curve shows the ABBA representation of the training data.}
\label{fig:M3}
\end{figure}

Leaving this concern about insufficient training data aside for the moment, it is still curious why the theta model~\cite{AN00, HB03}, a relatively simple statistical method, has been   found to outperform other comparably more complex and sophisticated models; see, e.g., \cite{MSA18}. 
Such comparative studies often use the popular sMAPE~\cite{A85} and MASE~\cite{HK06} distance measures when evaluating forecasting accuracy.  A closer inspection of the forecasts reveals that the theta method often produces (approximately) straight line forecasts, basically capturing just the trend of the time series. 
We illustrate this in Figure~\ref{fig:M3}, comparing the theta method with the raw LSTM and ABBA-LSTM models on the time series \texttt{N1500} from the M3 dataset.  
While the approximate straight line forecast may indeed yield  a small sMAPE score, it does not resemble the shape of the original time series. For a practitioner, the theta model forecast might be considered as ``unrealistic.'' Visually better forecasts are obtained by the raw LSTM and ABBA-LSTM models, with the ABBA-LSTM forecast  resembling the original data most closely.

\subsection{UCR Classification Archive}
The UCR Classification Archive \cite{UCRArchive} contains $128$ different classes of time series.  While the archive is not primarily intended for the purpose of forecasting, it provides a collection of time series with varying length from a good number of applications. We take the first time series from each class and z-normalise it, and keep only those time series which provide a training length of at least $100$ time series values for the raw LSTM model, and a string length of at least $20$  for the ABBA-LSTM model. We use the parameters $\texttt{tol} = 0.05$ and $\texttt{max\_k} = 10$ to obtain the ABBA representations~\cite{EG19b}. A total of $68$ time series are retained for this test.

The LSTM models contain two layers, each with $c = 50$ cells per layer, and are trained  with $50\%$ dropout rate. We train using a `stateful' procedure with lag parameter $\ell=10$ and a patience of $p=100$. Both models use the same untuned hyper parameters. Although the raw LSTM model has a total of $31051$ trainable parameters, and the ABBA-LSTM model has at most $33059$ trainable parameters (with the precise number depending on the alphabet used by ABBA, limited to at most nine symbols), the raw LSTM model has a much larger training set with at least $100-10=90$ time series values, whereas the ABBA-LSTM model has at least $20-10=10$ characters to train on.

After a model has been trained for a particular time series, we perform an iterated multi-step forecast to predict the next $k = 50$ time series values. We compare the similarity between the forecast and the ``truth'' using five similarity measures, including the sMAPE measure~\cite{A85}. The other four measures are Euclidean and dynamic time warping similarity measures on the original and differenced time series, respectively.

\begin{figure}[h!]
\centering
\includegraphics[scale=1]{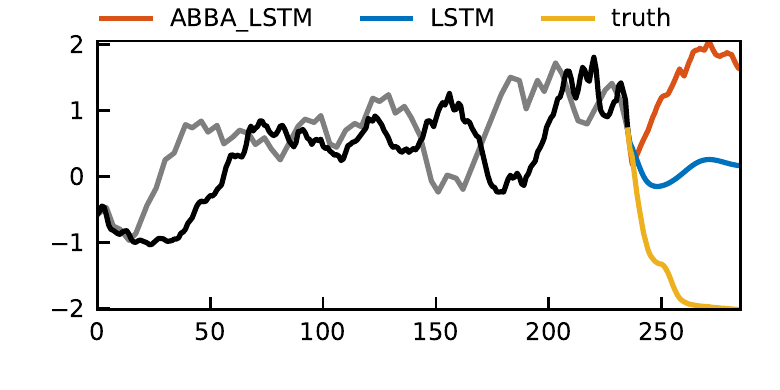}
\vspace*{-1mm}
\caption{Forecast of length $50$ for the \texttt{Coffee} time series from the UCR Classification Archive using the raw LSTM and ABBA-LSTM models. The grey line is  the ABBA representation of the training data.\label{fig:UCR_coffee}}
\end{figure} 

\begin{figure}[h!]
\centering
\includegraphics[scale=1]{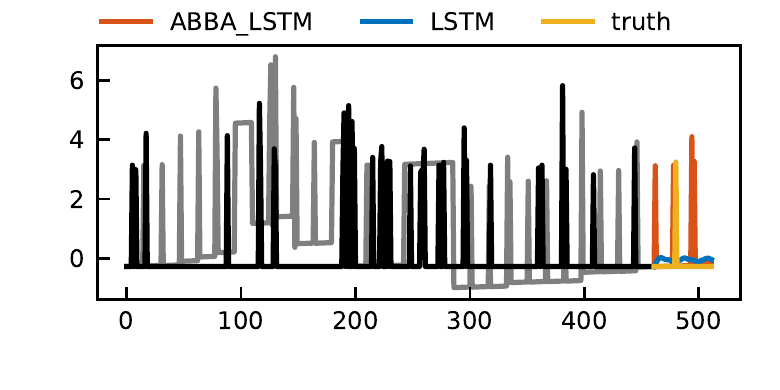}
\vspace*{-5mm}
\caption{Forecast of length $50$ for the \texttt{Earthquakes} time series from the UCR Classification Archive using the raw LSTM and ABBA-LSTM models. The grey line is the ABBA representation of the training data.\label{fig:UCR_earthquake}}
\end{figure} 

Examples of raw LSTM and ABBA-LSTM forecasts on two selected time series from the archive are shown in Figure~\ref{fig:UCR_coffee} and Figure~\ref{fig:UCR_earthquake}. The first example, Figure~\ref{fig:UCR_coffee}, illustrates a case where the ABBA-LSTM forecast is farther off the truth than the raw LSTM  model, but its forecast bears visually closer  resemblance to the historical training data. The other example, Figure~\ref{fig:UCR_earthquake}, demonstrates that the ABBA-LSTM model is able to forecast the spiky behaviour of the time series while the raw LSTM model produces a near-constant prediction. 

\begin{figure}[h]
\vspace*{3mm}
\hspace*{-5mm}\includegraphics[scale=1.16]{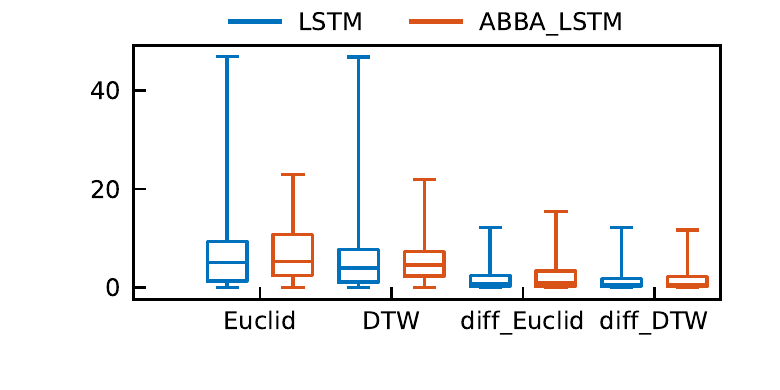}
\vspace*{-5mm}
\caption{Similarity measures between $50$ step forecasts and true values for the raw LSTM model and the ABBA-LSTM model.\label{fig:UCR_distance}}
\end{figure}

\begin{figure}[h!]
\hspace*{-3mm}\includegraphics[scale=1.14]{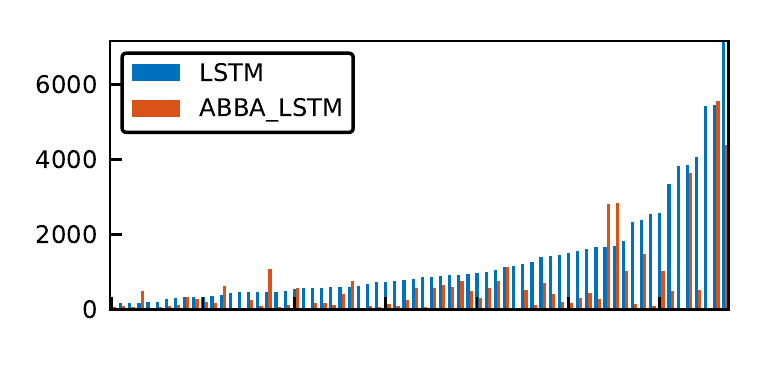}
\vspace*{-5mm}
\caption{Total training and forecasting runtimes of the raw LSTM and the ABBA-LSTM model for  each of the 68 time series.\label{fig:UCR_time}}
\end{figure}

Overall, using identical settings for the hyper parameters, both methods give comparable results in all four similarity measures, see Figure~\ref{fig:UCR_distance}. The raw LSTM model produces an average sMAPE score of $94.85$ and the ABBA-LSTM model produces an average sMAPE score of $88.39$ across all time series. A key advantage of the ABBA-LSTM model is the time reduction to build, train and forecast. On average, the raw LSTM model took $1293$ seconds per time series whereas the ABBA-LSTM model took $605$ seconds per time series. Figure~\ref{fig:UCR_time} compares the total runtime of both model types for each of the $68$ time series in the archive. In most cases, the ABBA-LSTM models are significantly faster to work with.

\section{Conclusion} \label{sec:con}
We proposed an approach to combine the effectiveness of machine learning methods for text generation and the ABBA symbolic representation to forecast time series. Many of the ideas discussed can be extended to other recurrent neural network models such as Gated Recurrent Units or the recent OpenAI GPT-2 framework~\cite{RW19}. Providing the time series is of sufficient length, the combined approach can lead to significant speed up of the training phase without degrading the forecast accuracy, whilst reducing the sensitivity to certain hyper parameters. 
Future research will be devoted to a more automatic way of specifying the number of LSTM cells and layers based on the complexity of the symbolic representation.


%

\ifCLASSOPTIONcompsoc
  \section*{Acknowledgments}
\else
  \section*{Acknowledgment}
\fi

This work was supported by the Engineering and Physical Sciences Research Council (EPRSC), grant EP/N509565/1. We  thank Sabisu and EPSRC for providing SE with a CASE PhD studentship. SG acknowledges support from the Alan Turing Institute under the EPSRC grant EP/N510129/1.

\ifCLASSOPTIONcaptionsoff
  \newpage
\fi



\bibliographystyle{IEEEtran}
\bibliography{references}
%



%

\begin{IEEEbiographynophoto}{Steven Elsworth}
received his MMath at The University of Manchester in 2016. He is currently a PhD student in Numerical Analysis at the University of Manchester. His research interests include rational Krylov methods and machine learning.
\end{IEEEbiographynophoto}
\begin{IEEEbiographynophoto}{Stefan G\"{u}ttel}
is Reader in Numerical Analysis at the University of Manchester. His work focuses on iterative methods for solving high-dimensional problems arising with differential equations and in data-driven applications, including the development of algorithms and software. He holds a Fellowship with the UK's Alan Turing Institute for data science and artificial intelligence.
\end{IEEEbiographynophoto}



\vfill


\end{document}